\PassOptionsToPackage{table,dvipsnames}{xcolor} 
\documentclass[10pt,twocolumn,letterpaper,hyphens]{article}

\usepackage{booktabs}       
\usepackage{amsfonts}       
\usepackage{nicefrac}       
\usepackage{microtype}      
\usepackage{calc}
\usepackage{graphicx}
\usepackage{caption}
\usepackage{subcaption}
\usepackage{booktabs}
\usepackage{multirow}
\usepackage{tikz}
\usepackage{colortbl}
\usepackage{graphicx}
\usepackage{siunitx}
\usepackage{multirow}
\usepackage{etoolbox}
\usepackage{fp}
\usepackage{amsmath}

\usepackage[pagenumbers]{iccv}  

\usepackage{pgfplots}
\pgfplotsset{compat=1.18}
\usepackage{textcomp}
\usepackage{graphicx}
\usepackage{siunitx}
\usepackage{enumitem}
\usepackage{multirow}
\usepackage{fp}
\usepackage{amsmath}
\usepackage{amssymb}
\usepackage{etoolbox}
\usepackage{algorithm}
\usepackage{algorithmicx}
\usepackage{algpseudocode}
\usepackage{wrapfig}
\usepackage{gensymb}
\usepackage{calc}
\usepackage{booktabs}       
\usepackage{amsfonts}       
\usepackage{wrapfig}
\usepackage{graphicx}
\usepackage{caption}
\usepackage{subcaption}
\usepackage{booktabs}
\usepackage{multirow}
\usepackage{colortbl}
\usepackage{float}
\usepackage[super]{nth}
\usepackage{tikz}
\usepackage{pgfplots}
\usepackage{colortbl}
\definecolor{wfx}{RGB}{150, 150, 0}

\definecolor{wd}{RGB}{128, 0, 255}
\definecolor{iccvblue}{rgb}{0.21,0.49,0.74}
\usepackage[pagebackref,breaklinks,colorlinks,allcolors=iccvblue]{hyperref}

\title{Harnessing Massive Satellite Imagery with Efficient Masked Image Modeling}

\author{%
\textbf{  Fengxiang Wang\textsuperscript{1},
  Hongzhen Wang\textsuperscript{2}\thanks{Corresponding authors} , 
  Di Wang\textsuperscript{3,}\textsuperscript{4}, 
  Zonghao Guo\textsuperscript{2}, 
  Zhenyu Zhong\textsuperscript{5},} \\
\textbf{Long Lan\textsuperscript{1}\footnotemark[1],
Wenjing Yang \textsuperscript{1}\footnotemark[1], 
Jing Zhang\textsuperscript{3}\footnotemark[1]}
  \\
  \textsuperscript{1} College of Computer Science and Technology, National University of Defense Technology \\
  \textsuperscript{2} Tsinghua University  
  \textsuperscript{3} School of Computer Science, Wuhan University \\
   \textsuperscript{4} Zhongguancun Academy
   \textsuperscript{5} Nankai University \\
}

\begin{document}
\maketitle
\begin{abstract}
Masked Image Modeling (MIM) has become an essential method for building foundational visual models in remote sensing (RS). However, the limitations in size and diversity of existing RS datasets restrict the ability of MIM methods to learn generalizable representations. Additionally, conventional MIM techniques, which require reconstructing all tokens, introduce unnecessary computational overhead. To address these issues, we present a new pre-training pipeline for RS models, featuring the creation of a large-scale RS dataset and an efficient MIM approach. We curated a high-quality dataset named \textbf{OpticalRS-13M} by collecting publicly available RS datasets and processing them through exclusion, slicing, and deduplication. OpticalRS-13M comprises 13 million optical images covering various RS tasks, such as object detection and pixel segmentation. To enhance efficiency, we propose \textbf{SelectiveMAE}, a pre-training method that dynamically encodes and reconstructs semantically rich patch tokens, thereby reducing the inefficiencies of traditional MIM models caused by redundant background pixels in RS images. Extensive experiments show that OpticalRS-13M significantly improves classification, detection, and segmentation performance, while SelectiveMAE increases training efficiency over 2$\times$ times. This highlights the effectiveness and scalability of our pipeline in developing RS foundational models.
The dataset, source code, and trained models will be released at \href{https://github.com/MiliLab/SelectiveMAE}{\color{magenta}SelectiveMAE}.

\end{abstract}    
\section{Introduction}
\label{sec:intro}
\begin{figure}[htbp]
\scriptsize
\centering
\includegraphics[width=\linewidth]{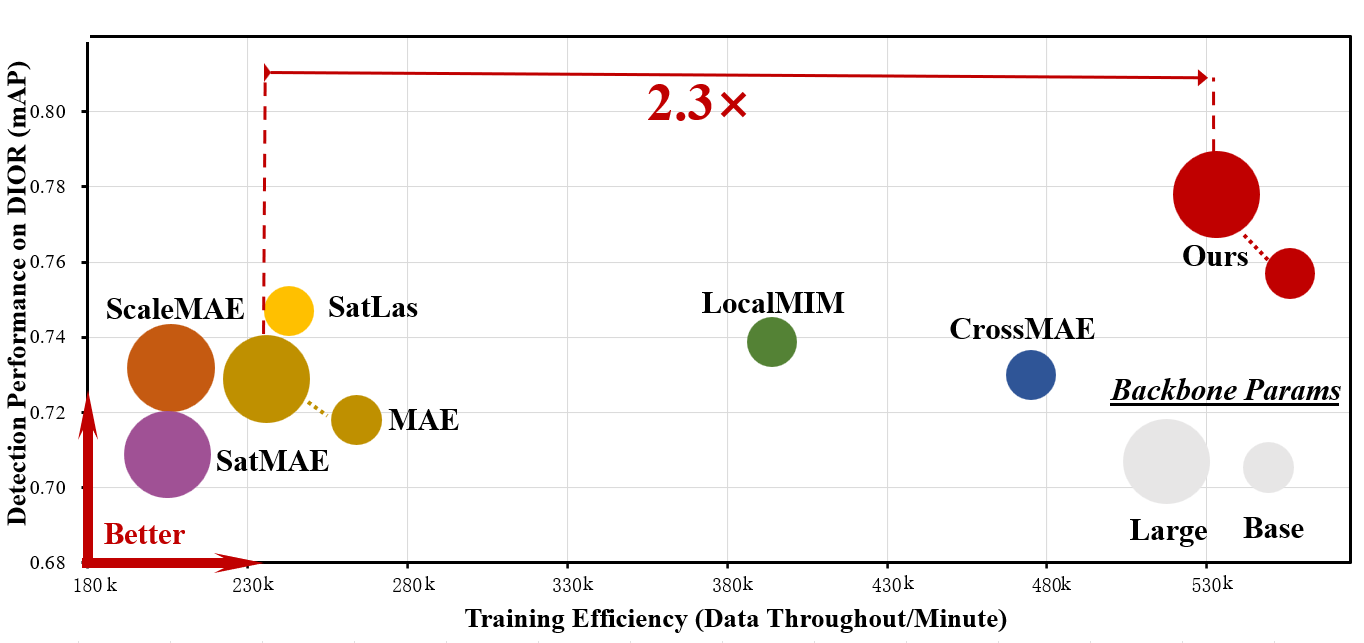}
\caption{Advantages of SelectiveMAE on finetuning performance and pre-training efficiency. The accuracy is evaluated on the DIOR detection dataset \cite{dior} with different versions of ViT \cite{vit}. The training efficiency is measured by \textit{Data Throughput/Minute}, \ie, the processed images per minute, on 8 $\times$NVIDIA A100 GPUs.}
\label{fig:abstract}
\vspace{-4mm}
\end{figure}
Over the past decade, advances in remote sensing (RS) technology and data acquisition have enhanced applications in ecosystem monitoring~\cite{appliaction1}, natural disaster management~\cite{appliaction2}, among others~\cite{my1,my2}. These rely on essential capabilities such as scene classification~\cite{classification1,classification2}, object detection~\cite{detection1}, change detection~\cite{detection2}, and semantic segmentation~\cite{segmentation}. However, each task typically demands substantial computational resources to train specialized models and learn task-specific feature representations.

\begin{figure*}[htbp]
\scriptsize
\centering
\includegraphics[width=\linewidth]{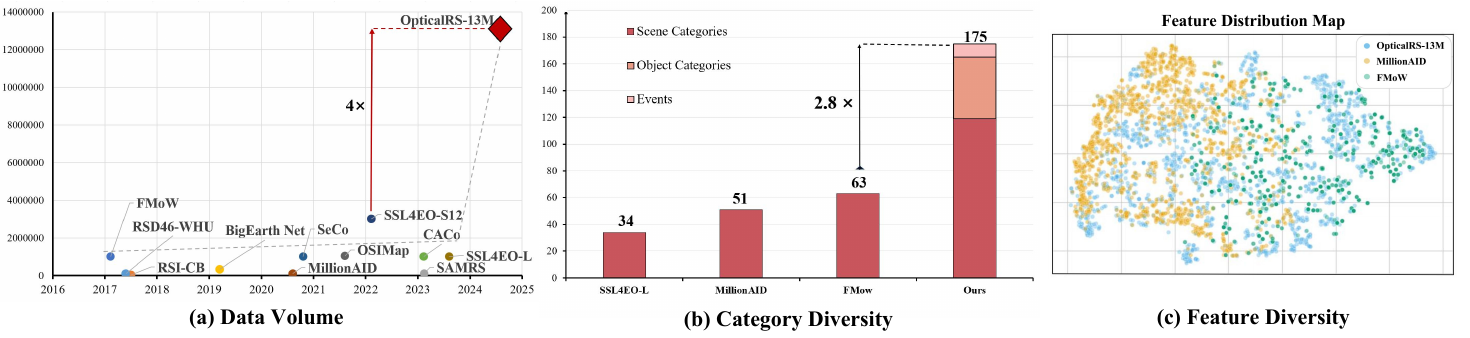}
\caption{Advantages of OpticalRS-13M in data volumn and diversity. (a) The volume of OpticalRS-13M surpasses at least 4$\times$ compared to existing RS datasets. (b) OpticalRS-13M includes significantly more sample categories than others, spanning both object and event types. y-axis denotes the data volume in (a-b). (c) OpticalRS-13M provides more diverse feature patterns than other large-scale RS datasets. Here, 1,000 samples are separately selected from different datasets, and their features from a pre-trained ViT-B are visualized using t-SNE~\cite{tsne}.}
\label{fig:dataset-abstract}
\vspace{-3mm}
\end{figure*}

Recent advances in self-supervised learning, particularly Masked Image Modeling (MIM) techniques~\cite{MAE,SimMIM}, have significantly improved the pre-training of visual foundation models~\cite{xu2022vitpose,zhang2023vitaev2,xu2023vitpose++,zhang2024vision,lan1,lan4}. This progress has led to the emergence of remote sensing foundation models (RSFMs), which provide general feature representations and achieve state-of-the-art performance across various remote sensing tasks~\cite{RVSA}. However, RSFMs still face two key challenges: (i) Compared to ImageNet-21k~\cite{imagenet21k}, existing RS datasets~\cite{MillionAID,CACo,SeCo,BigEarthNet} are significantly smaller (approximately 1 million vs. 14 million samples), limiting the effective MIM training of large models; and (ii) these datasets primarily capture global scene semantics~\cite{MillionAID,CACo,SeCo} but lack the diversity and fine-grained details essential for downstream tasks, hindering the generalization of learned representations.

To address these challenges, taking an example of visible light RS imagery, we propose a new pipeline to collect, create, and efficiently process a large RS dataset. Firstly, we reviewed publicly available remote sensing datasets from the past decade and selected them based on the DiRS (Diversity, Richness, and Scalability) principle \cite{MillionAID}. Nevertheless, issues such as inconsistent data sources, excessively large image sizes, and redundant pixels still exist. Therefore, we also apply exclusion, slicing, and deduplication processes to further improve data quality. Correspondingly, we obtain a large-scale RS dataset called \textbf{OpticalRS-13M} comprising 13 million images, which is designed to fully leverage the representation learning capabilities of MIM methods in RS applications. OpticalRS-13M exceeds previous RS datasets~\cite{GASSL,CACo,GFM,GeRSP,SAMRS,TOV,BFM,SatMAE,RingMo-Sense, lemevit,GeCo,ScaleMAE,MTP,Cross-ScaleMAE,SatMAE++,RVSA,RingMo,CMID}, being at least four times larger (Fig.~\ref{fig:dataset-abstract} left). Moreover, OpticalRS-13M encompasses a wide range of diverse RS scenarios encountered in downstream tasks such as object-level detection and pixel-level segmentation (Fig.~\ref{fig:dataset-abstract} right).

Despite substantial efforts in training RSFM using MIM methods \cite{GFM, oreole, SatMAE, ScaleMAE}, the computational burden and slow convergence when employing MIM training on large-scale RS datasets cannot be ignored. Specifically, pre-training on 1 million RS samples requires 107 hours for the ViT-B~\cite{vit} backbone on 8 Nvidia A100 GPUs~\cite{RVSA}. This issue becomes even more pronounced when training on larger datasets, \textit{e.g.}, OpticalRS-13M.
In natural scene analysis, this issue has led to numerous studies~\cite{SiameseMAE,LocalMIM,PCAE,EfficientMAE,fastmim,CrossMAE} aimed at improving MIM training efficiency. One approach is to accelerate the token reconstruction process by using decoders with fewer parameters~\cite{SiameseMAE,LocalMIM}. Another approach is to reduce the number of visible patch tokens input into the vision encoder~\cite{PCAE,EfficientMAE,fastmim}, speeding up feature extraction.

Conventional MIM approaches, following the encoding-then-decoding procedure, overlook the unique characteristics of RS images, which typically feature sparse foreground pixels and dense backgrounds \cite{RVSA,TOV}. This raises two key questions about how to efficiently conduct MIM training in the RS field: 1) Is it necessary to reconstruct all the redundant background patches during the MIM decoding process? 2) Is there a feasible way to encode fewer image patches (\textit{e.g.}, $\leq$25\%) to accelerate the convergence of MIM training? To address the first question, a measure-based selection process is needed to identify the appropriate patches for reconstruction. For the second question, the intuition is that the patch tokens used in the encoding-then-decoding procedure should effectively capture feature dependencies in RS images.

Regarding the above issues, in the second part of the pipeline, we introduce an MIM method called SelectiveMAE for efficiently processing RS images, which dynamically encodes and reconstructs patch tokens based on their semantic richness. Specifically, SelectiveMAE utilizes the Histogram of Oriented Gradients (HOG) algorithm to quantify the semantic richness of patches. Then, it selects a subset of patch tokens (\textit{e.g.}, $\leq$50\%) with higher HOG values for feature encoding (\textit{e.g.}, $\leq$15\%) and pixel reconstruction (\textit{e.g.}, $\leq$35\%). However, using an extremely low ratio of visible patches during MIM training can lead to gradient explosion. To mitigate this, we designed a Progressive Semantic Token Selection (PSTS) module, which dynamically selects semantically relevant patch tokens during the entire training phase. In the beginning, SelectiveMAE encodes semantically rich tokens and reconstructs semantically similar ones to warm up the training process. As training advances, SelectiveMAE shifts to reconstructing high-semantic tokens from encoded lower-semantic ones to capture complementary semantic dependencies. This analogical-to-complementary strategy allows SelectiveMAE to efficiently and progressively learn robust representations of RS images while accelerating MIM convergence (Fig. \ref{fig:abstract}). Our experiments reveal that 40\% of RS image patches are sufficient to train a comparable model, offering new insights into MIM training on RS images.

In summary, our main contributions are as follows:
\begin{enumerate}  
\item We introduce a new pipeline to collect, create, and efficiently process a large optical RS dataset for developing RS foundation models. Using this pipeline, we create the OpticalRS-13M dataset, which is a large-scale RS dataset comprising 13 million optical images with diversified coverage scenarios. 
\item For this pipeline, we introduce SelectiveMAE, an efficient MIM method tailored for RS image pre-training. It significantly accelerates convergence and enhances representation learning compared to the original MIM approach. 
\item Experiment results demonstrate the effectiveness and scalability of the proposed pipeline. OpticalRS-13M significantly enhances the performance of RS foundation models in downstream tasks, while SelectiveMAE achieves over $2\times$ speedup in pre-training compared to MAE~\cite{MAE}.
\end{enumerate}
\section{Related Work}
\label{sec:related}
\textbf{Remote Sensing Datasets.}
In recent years, many RS datasets have been created for tasks such as scene classification~\cite{BigEarthNet,tree}, object detection~\cite{fmow,floating-detection,ship1}, and segmentation~\cite{road,loveda,spacenet}. The availability of free, unlabeled satellite images has led to the development of large-scale RS datasets. Some works combine various sensor data to create extensive datasets. The SEN12MS dataset \cite{sen12ms} consists of 180,662 triplets, each containing synthetic aperture radar (SAR) and multispectral Sentinel-2 image patches, and MODIS land cover maps. While SSL4EO-S12 \cite{SSL4EO-S12} , SSL4EO-L \cite{SSL4EO-L} and SatlasPretrain \cite{SatLas} contain millions of images, the major data type in these datasets are multispectral and SAR, with only a small fraction being RGB. There are also some large-scale RS visual-language datasets such as Skyscript \cite{skyscript} and RS5M \cite{rs5m}, but they primarily focus on multimodal tasks. Currently, there are also some datasets that focus solely on visible light RS images. MillionAID~\cite{MillionAID}, SeCo~\cite{SeCo} and CACo~\cite{CACo} provide nearly a million images of the same location over different times. However, these datasets primarily target scene classification and often overlook fine-grained target information, limiting their utility for various downstream tasks. To address this gap, we introduce the OpticalRS-13M dataset, which is larger and more diverse. Using this dataset for pre-training, it significantly enhances performance across multiple downstream tasks.

\textbf{Masked Image Modeling.}
Inspired by the success of Masked Language Modeling (MLM) in NLP \cite{NLP,cai2021learning}, MIM has been developed for visual pre-training \cite{MIM1,SimMIM,MAE,lan2,lan3,ji2024segment,zhao2024parsing,wang2023hard,wang2025bootstrap}. MIM learns image representations by reconstructing masked tokens, focusing on various regression targets \cite{piexelnormal,Beit,HOGmask,deepfeature,frequency}, masking strategies \cite{ADIOS,SemMAE}, and reconstruction methods \cite{Pixeltransformer,DiffMAE,MST,AttMask}.
A major challenge for MIM is its high computational demand and lengthy pre-training times. To mitigate this, approaches such as asymmetric encoder-decoder strategies \cite{SiameseMAE, LocalMIM}, reducing input patches \cite{PCAE, fastmim}, or the Difficulty-Flatten Loss \cite{EfficientMAE} have been proposed. Additionally, CrossMAE \cite{CrossMAE} employs cross-attention between masked and visible tokens to enhance efficiency without sacrificing performance. However, these methods do not account for the unique characteristics of RS images, such as sparse foreground information and complex backgrounds.
Considering these issues, we introduce an efficient MIM method named SelectiveMAE that significantly speeds up pre-training, enhancing the practicality of developing RS foundation models based on large-scale datasets.

\textbf{Remote Sensing Foundation Models.}
Despite the abundance of RS data, much of it remains unlabeled and thus inaccessible for supervised learning~\cite{rsp}. Self-supervised learning methods have recently been employed to extract representations from unlabeled RS data. Due to the inefficiency of designing pretext tasks and gathering required data for contrastive self-supervised methods~\cite{GASSL,CACo,SeCo,DINO-MC,roma}, recent advancements have primarily centered around generative self-supervised methods, especially in MIM. Specifically, many studies aim to improve generative self-supervised algorithms by leveraging general image knowledge \cite{GFM, GeRSP, TOV}, scaling up parameter sizes \cite{BFM, oreole}, integrating spatio-temporal information \cite{SatMAE, RingMo-Sense, GeCo}, encompassing geometric attributes \cite{RVSA, MA3E}, handling multi-sensor data \cite{USat, SkySense, msGFM, FoMo-Bench, SpectralGPT}, and employing multi-scale concepts \cite{ScaleMAE, Cross-ScaleMAE, SatMAE++}. However, these methods have not effectively addressed the substantial computational burden associated with self-supervised pre-training in RS. In the paper, we propose a new pipeline that can collect, create, and efficiently process large amounts of optical RS data for developing RSFMs, enhancing the practicality of MIM pre-training on large-scale datasets.

\begin{figure}[tbp]

\centering
\includegraphics[width=\linewidth]{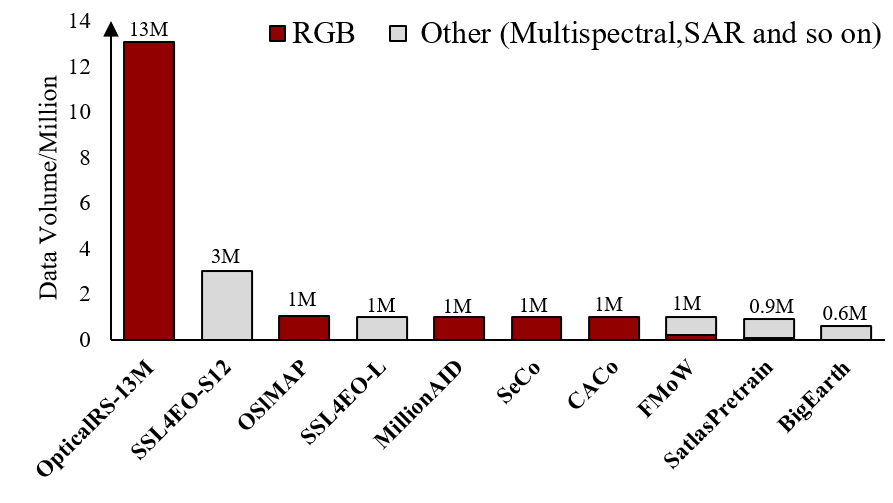}
\caption{The comparison between OpticalRS-13M and existing RS large-scale datasets in terms of data volume and image type.}
\label{fig:dataset-compare}
\vspace{-3mm}
\end{figure}
\section{Method}
\label{sec:method}
The proposed pipeline contains two critical components: dataset generation and efficient pre-training, which will be introduced in detail in the following text.
\vspace{-1mm}
\subsection{Dataset Curation}
\vspace{-1mm}
\label{section3-1}
Recent progress in self-supervised pre-training for RSFMs is limited by the relatively small scale and diversity of existing RS datasets compared to natural scene datasets. To overcome this issue, we curate a large-scale RS dataset with diverse coverage scenarios, named OpticalRS-13M. Detailed data collection and preprocessing are presented in the supplementary material.

\begin{figure*}[tbp]

\centering
\includegraphics[width=\linewidth]{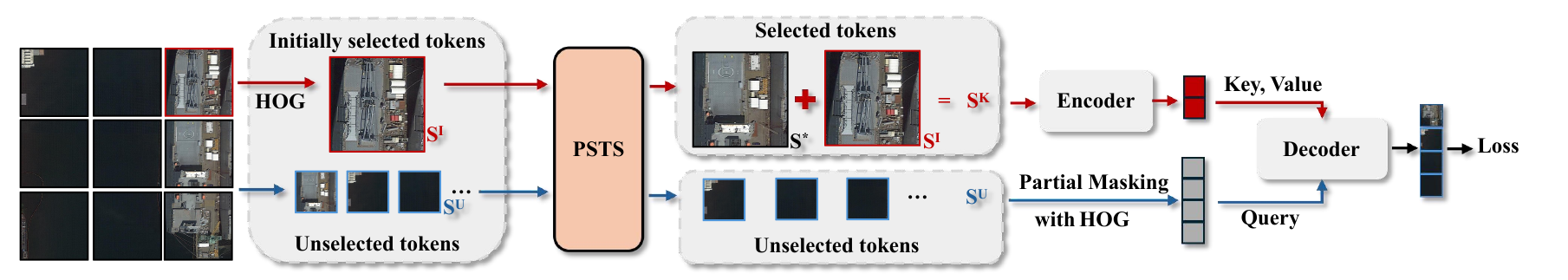}
\caption{Overview of SelectiveMAE. It inputs fewer visible patches and reconstructs only partial patches to accelerate training efficiency.}
\label{fig:main}
\vspace{-1mm}
\end{figure*}

\textbf{Data Volume.}
OpticalRS-13M comprises 13 million high-resolution visible light RS images, establishing a significant scale advantage over existing datasets, as illustrated in Figure \ref{fig:dataset-compare}. 
OpticalRS-13M contains 13,203,698 images with a total of 2,630,362,174,503 pixels, averaging 199,214 pixels per image.
We also offer a high-quality, lightweight version: OpticalRS-4M.
OpticalRS-4M contains 3,920,829 images with a total of 688,459,799,359 pixels, averaging 175,590 pixels per image.
While large-scale datasets such as SSL4EO-S12~\cite{SSL4EO-S12} and SatLasPretrain~\cite{SatLas} incorporate diverse data formats, \eg, including multispectral imagery, OpticalRS-13M is uniquely focused on visible light data. The dataset is carefully curated from multi-sensor acquisitions, leveraging imagery from high-resolution satellite platforms such as WorldView~\cite{WorldView}, QuickBird~\cite{QuickBird}, GeoEye~\cite{GeoEye}, and others. This specialization positions OpticalRS-13M as the most extensive visible light RS image repository to date, offering unparalleled utility for training vision foundation models for RS applications.

\textbf{Dataset Diversity.}
To demonstrate the broad scope of scenes and target categories covered by our dataset, we conducted a statistical analysis accompanied by a detailed label explanation, as shown in Fig. \ref{fig:dataset-abstract} (b) and (c). The analysis reveals that OpticalRS-13M comprises 12 main categories, each containing numerous subclasses. Unlike existing self-supervised optical RS datasets, OpticalRS-13M not only offers more comprehensive scene and target information but also introduces ``Events'' as a distinct high-level category. This category includes labels such as  ``Fire'', ``Flood'', ``Landslide'', ``Post-Earthquake'', and so on. This expanded information enhances the versatility of OpticalRS-13M for a wider range of downstream tasks.

\subsection{Efficient Pre-training}
After obtaining a large-scale RS dataset, the next step is conduct the self-supervised pre-training. However, as the capacity of the dataset grows, vast computational and time costs are required for existing MIM approaches used in the RS community, \textit{e.g.}, MAE \cite{MAE}. To address this issue, our pipeline introduces an efficient MIM method, SelectiveMAE. In this section, we first review the preliminaries of MAE, then we present the details of SelectiveMAE.

\label{section3-3}
\subsubsection{Preliminaries of MAE}

\label{section3-3-1}

\textbf{1) Masking.} Similar to supervised training of a standard ViT, MAE divides the image into regular, non-overlapping patches. It then samples a subset of these patches and masks the remaining ones. Typically, the masking ratio is 75\%, meaning only 25\% of the patches are input to the encoder. This random sampling follows a uniform distribution according to the masking ratio.
\textbf{2) MAE Encoder.} The encoder is a  standard ViT applied only to the visible, unmasked patches. It linearly projects the patches, adds positional embeddings, and processes them through a series of transformer blocks. By operating on a smaller subset of patches, the encoder enables training of large models with reduced computational and memory requirements.
\textbf{3) MAE Decoder.} The encoded tokens and masked tokens are fed into the decoder, which comprises transformer blocks with self-attention layers. The masked tokens are shared, learnable tensors enhanced with positional embeddings. The decoder, utilized only during pre-training, generates the output predictions for those masked tokens.
\textbf{4) Reconstruction Target.}
MAE predicts the pixel values for each masked patch, with each element in the decoder output representing a patch's pixel value vector. The loss function computes the mean squared error between the reconstructed targets and original patches.
 
However, applying MAE for self-supervised pre-training results in considerable training costs in terms of time complexity, especially when working with large RS datasets. To address this, we observe that RS optical images often contain redundant information, which could be omitted during training to improve pre-training efficiency. Specifically, we address two issues: 
\textit{\textbf{1) Is it necessary to reconstruct all the masked patches given the redundancy in RS images?}} 
\textit{\textbf{2) Can the visible patches input to the MAE encoder be further compressed to enhance acceleration?}}

\subsubsection{Partial Reconstruction}
\label{section3-3-2}
For question 1, previous research \cite{CrossMAE} has shown that for general images, when MAE reconstructs 75\% of the patches to calculate the loss, a specially designed decoder doesn't need to fully reconstruct all remaining patches. In fact, reconstructing just 50\% or even 25\% of the patches can achieve similar performance and speed up training. However, for RS images, if we randomly sample patches and remove most for reconstruction, the reconstructed patches might not be semantically rich ones. Using only a random subset for reconstruction even degrades performance.

To address this issue, we propose selecting semantically rich patches for reconstruction instead of random selection. Specifically, given an input image $x \in \mathbb{R}^{H \times W \times C}$, it is reshaped into $N=(H \times W) / p^{2}$ non-overlapping patches $x^{p} \in \mathbb{R}^{N \times (p^{2} C)}$, where $p$ is the patch size, $(H, W)$ is the size of the input image, and $C$ is the number of channels. These patches $\{x_{i}^{p}\}_{i=1}^{N}$ are then linearly mapped to patch embeddings. To retain positional information, positional embeddings are added to the patches. We select a portion of the patches to input to the encoder based on the masking ratio $m \in [0,1]$ ($m=85\%$ by default), as detailed in Sec.~\ref{section3-3-1}. The remaining patches serve as reconstruction targets for the decoder. 

Unlike MAE's masking ratio, we introduce a new reconstruction ratio $r$, the proportion of pixels to be reconstructed, denoted as $r \in [0,m]$ ($r=25\%$ by default). We compute the HOG features $HOG(\cdot)$ of the remaining patches and select those with the high HOG feature values according to the reconstruction ratio $r$, rather than using all patches. The process can be formulated as:
\begin{equation}
token_R=\{{x}_{i}^{p}|i\in {top}_{\lfloor\ r\times N\rfloor}(\mathit{HOG}(\left\{x_{i}^{p}\right\}_{i=1}^{m\times N}))\},
\label{eq1}
\end{equation}
where $token_R$ denotes the selected mask tokens for reconstruction and $\text{top}_n(\cdot)$ denotes the index set of the selected top $n$ tokens. The decoder uses a lightweight design based on cross-attention following CrossMAE~\cite{CrossMAE}. Experimental results in supplementary material show that this partial reconstruction strategy significantly increases the training throughput without affecting the learned representations.

\subsubsection{Progressive Semantic Token Selection}
\label{section3-3-3}

\begin{figure}[tbp]
  \centering
  \begin{minipage}[t]{0.22\textwidth} 
    \centering 
    \includegraphics[width=0.95\linewidth]{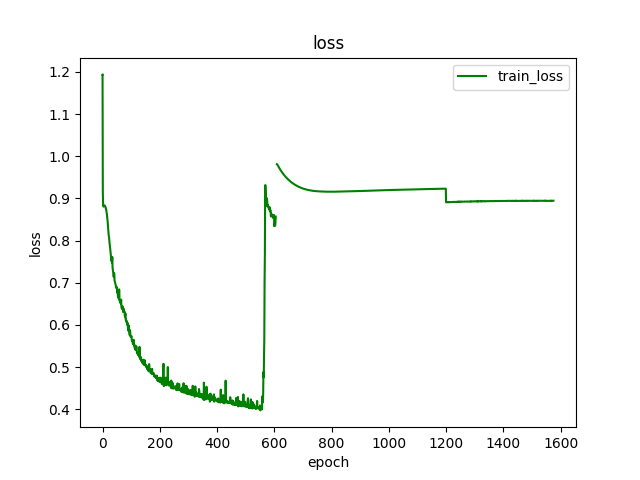}
  \end{minipage}%
  \begin{minipage}[t]{0.22\textwidth} 
    \centering
    \includegraphics[width=0.95\linewidth]{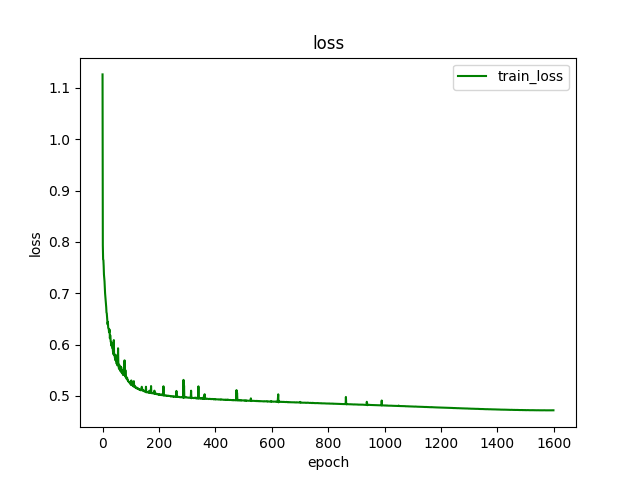}
  \end{minipage}
  \caption{Effectiveness of PSTS. Left: Using only 40\% of patches for encoding and reconstruction often leads to gradient explosions. Right: The training loss after adopting PSTS.}

  \label{fig-loss}
\end{figure}
For the second question, we initially tried a naive approach by increasing the masking ratio to 85\%, meaning only 15\% of the patches in each RS image were input to the encoder while keeping a 25\% reconstruction ratio as proposed in Sec.~\ref{section3-3-2}. However, during training, this setup often led to issues like gradient explosions or loss divergence, as shown in Fig.~\ref{fig-loss} left. The figure shows that using an extremely low portion of patches for encoding and reconstruction caused unstable training due to gradient explosions.

How can we achieve acceleration at a high masking ratio (e.g., 85\%) while ensuring MAE completes the pre-training task for RS images?
Inspired by curriculum learning ~\cite{curriculum_survey,curriculum,curriculum2}, which follows the principle of learning from easy to hard, we introduce the Progressive Semantic Token Selection (PSTS) module for patch selection, as  depicted in Fig.~\ref{fig:main}. In this module, we begin by selecting a limited number of patches and then select additional patches based on the similarity measure in the training epoch, dynamically transitioning from easily learned, semantically similar patches to more challenging, complementary ones.

For initialization, we employ a HOG selection strategy to choose the initial patch set from $S^N=\left\{x_{i}^{p}\right\}_{i=1}^{N}$, with a proportion $s \in [0,(1-m)/2]$. $m$ is the mask ratio. We define the initial set of selected token as:
\begin{equation}
\mit{S}^{I}=\{{S}^N(i)|i\in {top}_{\lfloor\ s\times N\rfloor}(\mathit{HOG}(S^{N} )\}.
\label{eq2}
\end{equation}
We select $\lfloor\ s\times N\rfloor$ tokens with the maximum HOG feature values (\textit{i.e.}, ${top}_{\lfloor\ s\times N\rfloor}$) from the original token set to form the initial token set. This simple yet effective strategy ensures that semantically rich tokens are selected.

After token initialization, we incrementally increase the number of tokens to guide training from easier to more challenging examples, while maintaining the final masking ratio for selected tokens, as outlined in Algorithm~\ref{algorithm1}. Given the high HOG feature values of the initial token sets, nearby tokens selected by PSTS also exhibit high HOG values. In partial reconstruction method in Section \ref{section3-3-2}, we filter unselected tokens to retain those with high HOG values as reconstruction targets. Thus, selecting nearby tokens brings encoding tokens and reconstruction targets closer, streamlining training. In contrast, if a token's feature distribution differs significantly from others, we consider these more challenging patches to learn.
\begin{figure}[t]
\vspace{-1mm}
\centering
\includegraphics[width=1.0\linewidth]{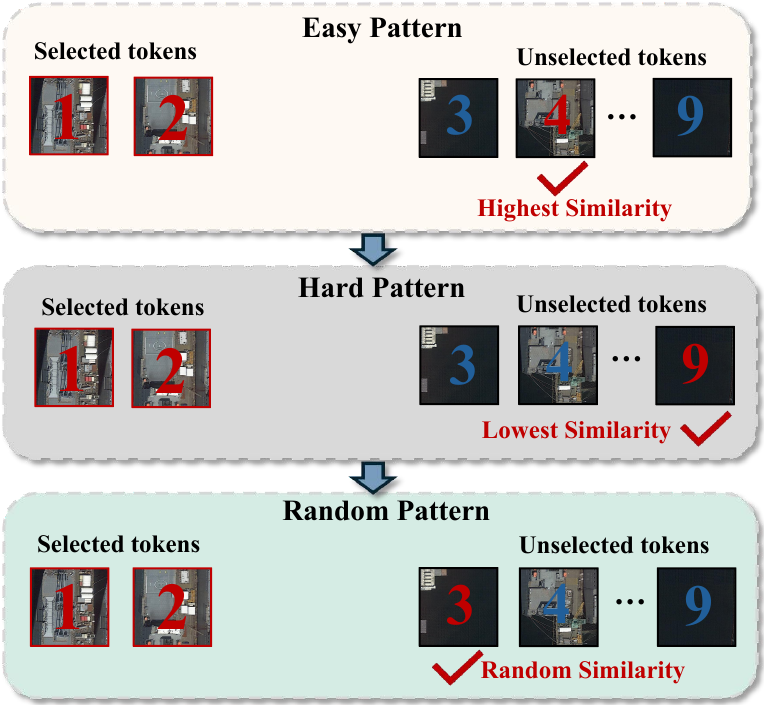}
\caption{Illustration of the patch selection process in PSTS.}
\label{fig:main2}

\end{figure}

\begin{algorithm}[t]
\scriptsize
\small
    \caption{Progressive Semantic Token Selection}
    \label{algorithm1}
    \begin{algorithmic}[1]
        \Require
        Number of training epochs $T$ and total training stages $N_g$ (\textit{i.e.}, $T/N_g$ epochs for each stage), masking rate $m$, input dataset $\mathcal{X}$
        \Ensure Obtain the selected tokens set ${S^{K}}$ and update ${S^{U}}$ in each epoch
        \For{$t\leftarrow 1$ \textbf{to} $T$}
            \State Sample data sample from $\mathcal{X}$, feed-forwarded through the embedding to obtain the output token set ${S}_N$
            
            \State $s \leftarrow  \frac{1}{2} (1-m)$
            \State Obtain ${S}^I$ via Eq.~\eqref{eq2} and initialize ${S^{U}}$
            \State Obtain the current training stage $\zeta=\lceil N_g*t/T\rceil$
           \State Calculate  $\mathit{distance}({S^{U}}\rightarrow{S^{I}})_i$ for \(i \in \{1, \cdots, |S^{U}|\}\) via Eq.~\eqref{eq3} and Eq.~\eqref{eq4}
        \State Obtain $S^{*}$ and update $S^{K},S^{U}$ via Eq.~\eqref{eq5} and Eq.~\eqref{eq6}
        \EndFor
    \end{algorithmic}
\end{algorithm}
Specifically, we select tokens from \(S^{U}\) based on \(S^{I}\). First, we use $\mathbb{S^{I}} \in \mathbb{R}^{|S^{K}|\times d}$ and $\mathbb{S^{U}} \in \mathbb{R}^{|S^{U}|\times d}$ to denote the matrix representation of the initial token set $S^{I}$ and the unselected token set $S^{U}$, where \(|\cdot|\) represents the number of tokens and \(d\) the feature dimension after the embedding layer. We use Cosine Distance to measure the dissimilarity between the tokens in these two sets:

\begin{equation}
\mathcal{D}(\mathbf{S}^{U}, \mathbf{S}^{I}) = 1 - \cos \left \langle \mathbf{S}^{U}, \mathbf{S}^{I} \right \rangle 
        = 1 - \frac{\mathbf{S}^{U} (\mathbf{S}^{I})^{\mathrm{T}}}{\lVert \mathbf{S}^{U} \rVert \cdot \lVert \mathbf{S}^{I} \rVert}
\label{eq3}
\end{equation}
where \(\mathbf{1}\) is an all-one matrix. \(\mathcal{D}(S^{U}, S^{I}) \in \mathbb{R}^{|S^{U}|\times |S^{I}|}\) represents the pairwise distances between tokens in \(S^{U}\) and \(S^{I}\). Next, we define the distance between the tokens in \(S^{U}\) to the initial token set \(S^{K}\) based on the selection criteria in each training stage as follows:
{\small
\begin{equation}
    \mathit{distance}({S^{U}}\rightarrow{S^{I}})_i =\begin{cases}
-\text{min}_j(\mathcal{D}({S^{U}}, {S^{I}})_{i,j}), & \zeta =1\\
\text{max}_j(\mathcal{D}({S^{U}}, {S^{I}})_{i,j}),&  \zeta =2\\
\text{random}_j(\mathcal{D}({S^{U}}, {S^{I}})_{i,j}),&\text{ otherwise }\\
\end{cases}, 
    \label{eq4}
\end{equation}
}
where \(i \in \{1, \cdots, |S^{U}|\}\), \(j \in \{1, \cdots, |S^{I}|\}\), and \(\zeta\) represents the training stage depending on the number of epochs. Finally, we sample $\lfloor N \times (1-m-s) \rfloor$ tokens from ${S^{U}}$ and add them with ${S^{I}}$ to form ${S^{K}}$, which can be formulated as follows:
{\small
\begin{equation}
    {S}^{*}=\{{S^{U}}(i)|i \in top_{\lfloor N\times (1-m-s) \rfloor}(\mathit{distance}({S^{U}}\rightarrow{S^{I}})_i )\},
    \label{eq5}
\end{equation}
}
\begin{equation}
        {S^{K}}={S^{I}} \cup \mit{S}^{*}, \quad
        {S^{U}}={S^{U}} \setminus \mit{S}^{*},
    \label{eq6}
\end{equation}
where $\mit{S}^{*}$ represents the selected token from ${S^{U}}$. The operations in Eq.~\eqref{eq5} and Eq.~\eqref{eq6} are performed in each training epoch. This process is summarized in Algorithm~\ref{algorithm1}. Fig. \ref{fig:main2} further provides an example to illustrate which patches should be selected at different stages, helping to facilitate readers' understanding.

\section{Experiment}
\label{sec:experiment}

In this section, we performed comprehensive comparative experiments to evaluate the performance of the proposed method on various downstream tasks, including RS scene classification, object detection, and semantic segmentation. Then, we validated the effectiveness of the OpticalRS-13M dataset and the SelectiveMAE method through ablation experiments. We also conducted further experiments to test the scalability of the pipeline. All experiments are performed by fine-tuning. \textbf{The pre-training and fine-tuning settings of our methods can be found in the supplementary material.}

\begin{table*}[htbp]
\footnotesize
    \centering
    \caption{Performance comparison results of different models. ``TR" represents the ratio of training data to the entire dataset. The first and second scores are marked by \textbf{bold} and \textcolor{blue}{blue}, respectively.  $\dagger$ means pre-training on 4 million images sampled from OpticalRS-13, with the epoch is set to 800. The overall accuracy is adoped as the metric for the scene classification task.}

    \resizebox{\linewidth}{!}{
        \begin{tabular}{lccccccccc}
        \toprule
        \multirow{4}{*}{Model}     & \multirow{4}{*}{Backbone}     & \multirow{4}{*}{Params (M)}  & \multirow{2}{*}{Data }
            & \multicolumn{2}{c}{Scene}
            & \multicolumn{2}{c}{Object}
            & \multicolumn{2}{c}{Semantic}
        \\  
        & && \multirow{2}{*}{  Throughput}&\multicolumn{2}{c}{Classification}
            & \multicolumn{2}{c}{Detection}
            &\multicolumn{2}{c}{Segmentation} \\\cmidrule(lr){5-6} \cmidrule(lr){7-8} \cmidrule(lr){9-10}
         & & & /Minute &  AID \cite{aid}& RESISC-45 \cite{RESISC-45}  & DIOR \cite{dior}  & DIOR-R \cite{dior-r}  & LoveDA \cite{loveda}   & SpaceNetv1 \cite{spacenet}
        \\
        \cmidrule(lr){5-6} \cmidrule(lr){7-8} \cmidrule(lr){9-10} & & & & TR=20\%/50\% & TR=10\%/20\%  & mAP$_{50}$ & mAP$_{50}$ &  mIoU & mF1
        \\
     \midrule
            SeCo	    \cite{SeCo}	       &   ResNet-50 \cite{resnet} & 26& -	&	93.47/95.99	&	89.64/92.91     &   -       &   -    &	43.63	&	77.09      \\
            GASSL	    \cite{GASSL}	    &	ResNet-50 \cite{resnet} & 26& -	&	93.55/95.92	&	90.86/93.06	   &   67.40	&	65.65   &	48.76	&	78.51    \\
	    TOV	        \cite{TOV}	       &   ResNet-50 \cite{resnet} & 26& 	 -    &	95.16/97.09	&	90.97/93.79   &   70.16	&	66.33 	&	49.70	&	-     \\
	    CACo	    \cite{CACo}	      &   ResNet-50 \cite{resnet} & 26& -	 &	90.88/95.05	&	88.28/91.94	      &   66.91	&	64.10 &	48.89	&	77.94    \\	     
      SatMAE	    \cite{SatMAE}	    &   ViT-L \cite{vit} 	&  307&  205k  &	95.02/96.94	&	91.72/94.10   &   70.89	&	65.66		&	-	    &	78.07    \\
	    ScaleMAE	\cite{ScaleMAE}	    &   ViT-L \cite{vit}  &  307&  	206k   &	96.44/97.58	&	92.63/95.04	  &   73.81	&	66.47   &	-	    &	-      \\
	    SSL4EO	    \cite{SSL4EO-S12}	    &   ViT-S \cite{vit}  & 22& -   &	91.06/94.74	&	87.60/91.27	   &   64.82	&	61.23	 &	-	    &	-    \\
	    RingMo	    \cite{RingMo}	    &   Swin-B \cite{swin}	& 88&   -   &	96.90/98.34	&	94.25/95.67	  &   75.90	&	-	 &	-	&	-      \\
	    SatLas	    \cite{SatLas}	    &   Swin-B \cite{swin}	& 88 &  243k  &	94.96/97.38	&	92.16/94.70    &   74.10	&	67.59	 &	-  	&	-   \\
	    GFM	        \cite{GFM}	       &   Swin-B \cite{swin}	& 88 &  -  &	95.47/97.09	&	92.73/94.64	  &   72.84	&	67.67	 &	-   	&	-     \\ 
     RVSA	    \cite{RVSA}	     &   ViT-B+RVSA \cite{RVSA} & 	86 &   -    &97.03/98.50	&	93.93/95.69	       &   75.80	&	68.06	&	51.95	&	-      \\

     OREOLE \cite{oreole} & ViT-G \cite{oreole} & 914 & - & 96.71/$\quad$-$\quad$ & $\quad$-$\quad$/$\quad$-$\quad$ &77.40 & \textcolor{blue}{71.31} & \textcolor{blue}{54.00} & - \\

     \midrule
     MAE \cite{MAE}$\dagger$ &  ViT-B \cite{vit}  & 86 & 264k &96.58/98.02  &    92.44/94.43  &  75.40 &  67.35&  52.80& 79.41   \\
        \midrule
        SelectiveMAE $\dagger$       & ViT-B \cite{vit}  & 86 & 556k & 96.90/98.12      & 93.35/94.58   & 75.70     & 67.78 & 53.05     & \textbf{79.50}\\ 
        SelectiveMAE$\dagger$            & ViT-L \cite{vit}  & 307 & 533k & \textcolor{blue}{97.25}/\textcolor{blue}{98.50}      & \textcolor{blue}{94.57}/\textcolor{blue}{95.77}  & \textcolor{blue}{77.80}      & 70.31 & \textbf{54.31}    & 79.46     \\
        \midrule
        SelectiveMAE             & ViT-B \cite{vit}  & 86 & 556k  & 97.10/98.28      & 93.70/95.48   &  75.80     &  67.69 &   52.68   &   79.44           \\ 
        SelectiveMAE            & ViT-L \cite{vit}  & 307 &  533k &\textbf{ 97.49}/\textbf{98.52 } & \textbf{94.73}/\textbf{96.36} &  \textbf{78.70}    &  \textbf{71.75} &   53.92   &   \textcolor{blue}{79.48}          \\ 
        \bottomrule
    \end{tabular}
}

\label{table-downstream}
\end{table*}

\subsection{Main Results}
\label{section4-1}

We compare SelectiveMAE to state-of-the-art RSFMs. In addition to benchmarking against MIM-based approaches such as SatMAE \cite{SatMAE} and ScaleMAE \cite{ScaleMAE}, we also compare with contrastive learning-based methods, including GASSL \cite{GASSL} and SeCo \cite{SeCo}, as well as advanced supervised learning models like SatLas \cite{SatLas}. Notably, these comparative methods were specifically designed for increasing performance rather than accelerating pre-training. \textbf{The Scene Classification and Detection comparison results are taken directly from SkySense~\cite{SkySense}, while those for Semantic Segmentation are exactly sourced from MTP~\cite{MTP}.} Although their speed performance was not reported, we expect them to lag behind the baseline MAE. The performance of ViT-B and ViT-L on various downstream tasks after pre-training is presented in Table \ref{table-downstream}. The results indicate that after pre-training on OpticalRS-13M, SelectiveMAE outperforms other RSFMs. Notably, it not only outperforms baseline (MAE) but also achieves this at 2.1 times the speed (556/264). This makes SelectiveMAE particularly well-suited for large-scale datasets, offering significant time savings during training on OpticalRS-13M.

\textbf{Scene Classification.} 
We first assess the pre-trained model’s performance on the scene classification task to evaluate its overall representational capability without additional decoders. We use two scene classification datasets: AID \cite{aid} and RESISC-45 \cite{RESISC-45}, following training details and train-test splits as in \cite{RVSA,ScaleMAE}. Table~\ref{table-downstream} shows that SelectiveMAE performs competitively against other pre-training methods on both datasets. 
Additionally, when scaled to ViT-L, our model outperforms OREOLE \cite{oreole} and other competitors, indicating the efficacy of SelectiveMAE pre-training on the OpticalRS-13M dataset, which enables strong feature representation learning while maintaining efficient scalability with increasing model size.

\textbf{Horizontal \& Oriented Object Detection.} 
We utilized the well-established DIOR dataset for horizontal object detection \cite{dior} and its improved variant DIOR-R for oriented object detection \cite{dior-r}. Following the methodologies of previous works \cite{RVSA,RingMo}, we maintained consistent experimental setups, employing Faster-RCNN \cite{faster_rcnn} and Oriented-RCNN \cite{oriented_rcnn} as detectors for each dataset. The results are summarized in Table~\ref{table-downstream}. Our approach, utilizing a ViT-B backbone, demonstrates competitive or superior performance compared to other methods with a Swin-B backbone, such as RingMo~\cite{RingMo}. When using the larger ViT-L backbone, SelectiveMAE shows enhanced performance across both detection datasets, even outperforming OREOLE \cite{oreole}, which has close to 1B parameters, underscoring the excellent scalability of our method.

\textbf{Semantic Segmentation.} 
We further evaluate the performance of the pre-trained model on pixel-level perception tasks, particularly semantic segmentation, using two well-known RS datasets: LoveDA \cite{loveda} and SpaceNetv1 \cite{spacenet}. Our implementation follows \cite{RVSA}, utilizing UperNet \cite{upernet} as the segmentation framework. Table~\ref{table-downstream} demonstrates the clear superiority of SelectiveMAE over its competitors in semantic segmentation tasks. SelectiveMAE focus on semantically rich patches, such as the boundaries between foreground objects and background stuff, leading to better representation learning for segmentation.

\subsection{Ablation Study}
To validate the efficacy of SelectiveMAE, we conducted ablation experiments about HOG Selecting, Similarity Measure and Selection Pattern. To further verify the pipeline's effectiveness, we conducted separate evaluations of both the OpticalRS-13M dataset generated by the pipeline and the accelerated performance of SelectiveMAE. Additionally, we compared the SelectiveMAE with the CrossMAE, a MAE-based method for natural images. More ablation experiment results are presented in the supplementary material.

\textbf{Alternatives of HOG Selecting.} To verify the efficacy of HOG in token selecting, we conduct two ablation experiments using MillionAID dataset \cite{MillionAID} to replace HOG: (1) we used a pretrained Swin-B \cite{swin} model and applied k-nearest neighbor clustering to the extracted features. (2) We employed a lightweight feature extractor network based on Adamae~\cite{adamae}. Results in the Table \ref{table-ablation-HOG} demonstrate HOG’s pre-training speed advantage. While using Swin-B slightly outperforms HOG, it is highly inefficient, and pre-training a foundation model like Swin requires massive data and computational resources. For RS images with abundant low-level features, HOG effectively selects tokens while ensuring substantially higher training speeds.

\begin{table}[t]

\centering
\scriptsize
    \caption{Ablation experiment results of HOG selection through pre-training on the MillionAID dataset for 800 epochs.}

    \resizebox{\linewidth}{!}{
    \begin{tabular}{ccccc}
   \toprule 
    \multirow{2}{*}{Method} & \multirow{2}{*}{Params}&  Data Throughput& AID  &  RESISC-45  \\\cmidrule(lr){4-5}
    & & / Minute & TR=20\%/50\% & TR=10\%/20\% \\
    \midrule
    
    Adamae~\cite{adamae} & 2.36M & 498k  & 88.78/91.25&   85.72/87.44 \\
   Swin-B & 88M & 356k   &93.21/96.48 &  89.94/93.72 \\
   \rowcolor{black!15} HOG& - &  556k  &  93.17/96.12   &  89.21/92.31  \\
   \bottomrule 
    \end{tabular}}
    \label{table-ablation-HOG}
\end{table}

\textbf{Ablation of Similarity Measure and Selection Pattern.} To comprehensively assess the PSTS, we tested different Similarity Measures and the Selection Pattern. The results in the Table~\ref{table-ablation-sim} reveal that different similarity measures had minimal impact on performance. However, changing the selection pattern led to a performance drop, confirming the effectiveness of the PSTS.
\begin{table}[t]
\centering
  \footnotesize
  \caption{ Ablation experiment results of different similarity measures and selection patterns in PSTS through pre-training on the MillionAID dataset for 800 epochs. }
  \setlength{\tabcolsep}{6pt}
    \resizebox{1.0\linewidth}{!}{\begin{tabular}{cccc}
   \toprule
    \multirow{2}{*}{Selection Pattern} & \multirow{2}{*}{Similarity Measure} &  AID & RESISC-45 \\\cmidrule(lr){3-4}
    &  &  TR=20\%/50\% & TR=10\%/20\%\\
    \midrule
     near-far-random   & Euclidean distance & 93.02/95.96     &  88.97/92.07  \\ 
    near-far-random &   Manhattan distance &  92.92/95.89  &  89.17/91.92  \\\midrule
    far-near-random & Cosine distance  & 90.12/92.78  & 85.81/88.80\\
       \rowcolor{black!15} near-far-random  & Cosine distance  &  93.17/96.12   &  89.21/92.31 \\
  \bottomrule
    \end{tabular}}
    \label{table-ablation-sim}
   \vspace{-2mm}
\end{table}

\begin{table}[htbp]
\centering
\scriptsize
\caption{
Performance comparison of using different pre-training settings on datasets, methods and epochs, where the ViT-B \cite{vit} is adopted as the network.}

 \resizebox{\linewidth}{!}{\begin{tabular}{lccccc}
\toprule
\multirow{2}{*}{Dataset} & \multirow{2}{*}{Method}  & \multirow{2}{*}{Image Number}  &\multirow{2}{*}{Epoch}    & AID \cite{aid}& RESISC-45 \cite{RESISC-45}  \\  \cmidrule(lr){5-6}
       &       &     &     & TR=20\%/50\%  & TR=10\%/20\% \\
\midrule
\multicolumn{6}{c}{\textit{\textbf{Different Datasets}}} \\
\midrule
MillionAID \cite{MillionAID} & MAE    &   1 million    & 800    & 94.92/97.38    & 89.20/93.60 \\
OpticalRS-13M   & MAE & 1 million    & 800    & \textbf{96.26}/\textbf{97.98} & \textbf{91.53}/\textbf{93.88}  \\ 
\midrule
\multicolumn{6}{c}{\textit{\textbf{Equivalent Data Throughput}}} \\
\midrule
OpticalRS-13M & MAE  &  2 million    & 400   &  96.64/98.10 & 91.80/94.31  \\
OpticalRS-13M & MAE &  3 million     & 267    & \textbf{96.67}/98.18 &  92.24/\textbf{94.41} \\
OpticalRS-13M   & MAE     & 4 million    & 200    &96.10/98.03    & \textbf{92.38}/94.30 \\
OpticalRS-13M   & MAE     & 8 million    & 100    &96.58/\textbf{98.26} & 91.83/93.99  \\
OpticalRS-13M   & MAE     & 13 million    & 67    &96.28/98.06 & 91.41/93.60 \\
\midrule
OpticalRS-13M   &SelectiveMAE   & 1 million  &800 &  96.29/97.78    &  91.41/93.48  \\
OpticalRS-13M   &SelectiveMAE &   4 million & 200 &  95.96/\textbf{98.06}  & \textbf{92.06/94.05}   \\
OpticalRS-13M   &SelectiveMAE &   13 million & 67 &  \textbf{96.31}/97.95 & 91.88/93.76 \\   
\midrule
\multicolumn{6}{c}{\textit{\textbf{More Epoch}}} \\
\midrule
OpticalRS-13M   &     MAE   & 13 million  &67 &  96.28/98.06    &  91.41/93.60\\
OpticalRS-13M   &    MAE &   13 million & 100 & 96.34/98.13  &  91.79/93.85 \\
OpticalRS-13M   &    MAE &   13 million & 200 & 96.51/98.22 & 92.19/94.26\\
\rowcolor{black!15} OpticalRS-13M   & MAE &   13 million & 800 & \textbf{97.10/98.28} & \textbf{93.70/95.48}\\
\bottomrule
\end{tabular}}
\label{Table-dataset1}
\end{table}

\textbf{Efficacy of OpticalRS-13M.}

As illustrated in Table~\ref{Table-dataset1}, when pre-training with a randomly sampled subset of 1 million images from OpticalRS-13M, the model achieves superior performance compared to MillionAID \cite{MillionAID}, underscoring the efficacy of OpticalRS-13M for representation learning.
To further assess dataset diversity, we conducted experiments with equivalent data throughput during pre-training. The results in Table~\ref{Table-dataset1} reveal that optimal performance is attained by varying training configurations rather than utilizing the entire 13 million image corpus, regardless of the pre-training methodology. In our opinion, this observation highlights the inherent diversity of OpticalRS-13M, as training with fewer epochs proved insufficient for the MIM method to fully exploit the dataset’s potential, leading to model underfitting.
To validate this hypothesis, we extended the training schedule, and the experimental results presented in the last part of Table~\ref{Table-dataset1} confirm our claim.

\begin{table}[tbp]
\scriptsize
\centering
\caption{The efficiency comparison between SelectiveMAE and the baseline method MAE \cite{MAE} across different backbones on the part of OpticalRS-13M dataset \cite{MillionAID} with 800 epochs. The memory is measured on a single NVIDIA A100 GPU with a batch size of 256.}

 \resizebox{\linewidth}{!}{
 \begin{tabular}{cccccc}
		\toprule
		Model &  Backbone & Data Volume  & Training Time (h) & GPU Memory (MB)  \\
		\midrule
		MAE  &  ViT-B & 1 million   & 51 & 17628 \\

    		\rowcolor{black!15} SelectiveMAE  &  ViT-B & 1 million  & \textbf{24 {\color{red}($2.1\times$, -27)}} & \textbf{9570 {\color{red}($1.8\times$)}} \\
            \midrule
             MAE &  ViT-L & 1 million  & 57 & 30530 \\
		\rowcolor{black!15} SelectiveMAE & ViT-L &   1 million & \textbf{25 {\color{red}($2.3\times$, -32)}} & \textbf{18790 {\color{red}($1.6\times$)}} \\
		\midrule
		MAE  &  ViT-B & 4 million   & 177 & 17628 \\
    		\rowcolor{black!15} SelectiveMAE  &  ViT-B & 4 million  & \textbf{86 {\color{red}($2.1\times$, -81)} } & \textbf{9570 {\color{red}($1.8\times$)}} \\
		\bottomrule
	\end{tabular}}
    \label{table-pretrain}
     \vspace{-1mm}
\end{table}

\textbf{Efficiency Advantage of SelectiveMAE.} 
To further highlight the efficiency advantage of SelectiveMAE, we evaluate the training time and memory footprint during pre-training of different backbones on the part of OpticalRS-13M dataset, as shown in Table \ref{table-pretrain} (the corresponding accuracies are presented in Table \ref{table-downstream}). It can be seen that, since only 40\% of the image patches (15\% for the encoder and 25\% for the decoder) are involved in the calculation, our methods present more than doubled training acceleration compared to the vanilla MAE baseline, where the acceleration advantage is more significant for larger models. It is worth noting that SelectiveMAE also reduces memory footprint by using fewer patches.
To further evaluate the scalability of our pipeline, we organized datasets containing 4 million samples and conducted experiments. The results in Table~\ref{table-pretrain} show that,
when the dataset was expanded to 4 million samples, the time savings increased from 27 (51-24) to 81 hours. These findings demonstrate that our pipeline scales effectively with larger datasets, offering significant pre-training acceleration.

\textbf{Comparison with CrossMAE~\cite{CrossMAE}.}
CrossMAE, an MAE-based self-supervised pretraining method, is originally designed for natural images. Since SelectiveMAE adopts the same decoder as CrossMAE \cite{CrossMAE} during pre-training, and incorporates a HOG-based strategy tailored to RS images inspired by its partial reconstruction concept, as detailed in Section 3.2.2. In this section, we have also complemented related comparison experiments. Nevertheless, due to CrossMAE has not been pretrained in the RS field, we conducted a fair evaluation through pretraining it on the OpticalRS-13M dataset using the official code. Table \ref{table-ablation-crossmae} shows that SelectiveMAE outperforms CrossMAE in both accuracy and efficiency. By integrating RS-specific enhancements, PSTS and partial reconstruction, SelectiveMAE proves to be a highly effective self-supervised learning approach for RS.

\begin{table}[t]
\centering
\setlength{\tabcolsep}{6pt}
  \footnotesize
  \caption{Performance comparison between our method and CrossMAE, where 400 million images of OpticalRS-13M are randomly sampled for pre-training 800 epochs. }
    \resizebox{1.0\linewidth}{!}{\begin{tabular}{cccc}
   \toprule
    \multirow{2}{*}{Method} & \multirow{2}{*}{Data Throughput / Minute} & AID  & RESISC-45 \\\cmidrule(lr){3-4}
    &  & TR=20\%/50\% & TR=10\%/20\%\\
    \midrule
   
    CrossMAE(ViT-B)  & 475k  &   96.12/97.18  &92.08/94.12 \\
     \rowcolor{black!15} SelectiveMAE(ViT-B) & 556k  &  96.78/98.12   & 93.35/94.58\\
   \bottomrule
    \end{tabular}}
    \label{table-ablation-crossmae}
    \vspace{-4mm}
\end{table}

\section{Conclusion}
\label{sec:conclusion}
In this paper, we introduce a new pre-training pipeline for RS models, featuring the creation of a large-scale RS dataset and an efficient MIM approach. We first curated OpticalRS-13M, a large-scale optical remote sensing dataset for unsupervised learning. Unlike previous RS datasets, OpticalRS-13M offers a larger and more diverse image set with fine-grained details relevant to downstream tasks. Benchmarking representative MIM methods on OpticalRS-13M highlights its advantages in these tasks. Then, we present SelectiveMAE to reduce the computational overhead of MIM training on large-scale RS datasets. This efficient MIM method dynamically encodes and reconstructs tokens based on their semantic richness. SelectiveMAE significantly accelerates training, demonstrating that using only 40\% RS image patches is sufficient for training a comparable MIM model. Extensive experiments show that OpticalRS-13M significantly contributes to improving classification, detection, and segmentation performance, while SelectiveMAE achieves over a $2\times$ speedup along with GPU memory savings, highlighting the effectiveness and scalability of our pipeline in developing RS foundational models.
\section{Acknowledgements}
We gratefully acknowledge Prof. Zhiyuan Liu and Prof. Maosong Sun for their helpful discussions and support during this research. This work was partially supported by the National Natural
Science Foundation of China (No. 62372459, No.62376282 and No. 624B2109) and the Major Special Project of China Innovation Challenge (Ningbo) under Grant 2024T008.

{
    \small
    \bibliographystyle{unsrtnat}
    \bibliography{main}
}

\clearpage
\section{Supplementary Material}
\subsection{Overview}
This material provides additional details of the proposed \textbf{SelectiveMAE} and \textbf{OpticalRS-13M}, as well as experimental results that are omitted from the main body of this paper due to the page limit, which are organized as follows:
\begin{itemize}
    \item Sec.~\ref{app-sec1} offers more details of the OptimalRS-13M.
    \item Sec.~\ref{app-sec2} provides a detailed ablation study of SelectiveMAE.
    \item Sec.~\ref{app-sec3} provides the full experiment configurations of pretraining and downstream tasks.
    \item Sec.~\ref{app-vis} visualizes the category structures and samples of OpticalRS-13M and predicted results of SelectiveMAE on the downstream tasks.
    \item Sec.~\ref{app-datasheets} offers the Datasheets for the OpticalRS-13M dataset.
    \item Sec.~\ref{app-limitation} offers the Limitation and Potential Societal Impact.
    
\end{itemize}

\subsection{More details on OptimalRS-13M.}
\label{app-sec1}

\textbf{Data Collection}
We began by reviewing publicly available RS datasets from the past decade. Recognizing that not all data are suitable for self-supervised pre-training, we adopted specific criteria for data collection, as outlined in DiRS \cite{MillionAID}: \textbf{1) Diversity}: A dataset is diverse if its images capture various typical visual features of relevant scenes or targets and offer complementarity. \textbf{Intra-class Diversity}: High within-class diversity ensures comprehensive representation of real-world target distributions. We include data from different sources of the same scene or target, incorporating change detection data to capture object variations by geographic location and imaging time. \textbf{Inter-class Diversity}: To enable self-supervised pretraining to distinguish categories effectively, we included as many fine-grained categories as possible, especially those with high semantic overlap, helping the model learn robust feature representations. \textbf{2) Richness}: Beyond diversity, richness is essential. We ensure varied content features and large sample sizes by collecting images under multiple conditions (e.g., weather, seasons, lighting), introducing variability in translation, viewpoint, object posture, spatial resolution, background, and occlusion. Sampling across imaging conditions provides a comprehensive real-world representation, enhancing model's representation and generalization  capabilities. \textbf{3) Scalability}: Scalability is crucial given the evolving applications of remote sensing images. We continuously expand the dataset based on scene and target categories across classification, detection, and segmentation tasks.
In summary, following the guidelines, we constructed a large-scale RS image dataset with diverse coverage scenarios.

\textbf{Data Preprocessing}
Following data collection, challenges such as inconsistent data sources, oversized images, and redundant pixels remained. To address these, we implemented a standardized, scalable data preprocessing workflow consisting of four steps: 1) We focused exclusively on visible light images for this study, excluding multispectral and SAR data, though future updates will incorporate these modalities for multi-modal self-supervised pre-training. 2) To manage the large image sizes, we randomly cropped high-resolution images into smaller slices. We divided the high-resolution images into sub-images with sizes varying between 64 $\times$ 64 and 1,024 $\times$ 1024 pixels. 4) The remaining images were combined and duplicates were removed using a two-phase approach: first, a coarse phase with perceptual hashing~\cite{hash}, and then a refined phase involving manual review. This ensured that only highly similar images were excluded.

\subsection{Ablation study of SelectiveMAE}
\label{app-sec2}
We conducted ablation experiments to determine the best design choice for SelectiveMAE. The results of these experiments are presented in Table \ref{table-ablation1}-\ref{table-ablation2}.

In fact, reconstructing just 50\% or even 25\% of the patches can achieve similar performance and speed up training. However, for RS images, if we randomly sample patches and remove most for reconstruction, the reconstructed patches might not be semantically rich ones. As shown in Table~\ref{table-ablation1}, using only a random subset for reconstruction degrades performance.

ATS, introduced by AdaMAE~\cite{adamae}, is an adaptive masking strategy for video domains that uses a learnable auxiliary sampling network to select visible tokens based on semantic context. By estimating a categorical distribution over spacetime patch tokens, it prioritizes high spatiotemporal information regions, enabling 95\% token masking to reduce memory usage and accelerate pretraining. However, despite faster training, the auxiliary network severely compromises effectiveness to an unacceptable level.

Based on Table~\ref{table-ablation1}, we ultimately chose the HOG-based selective reconstruction strategy, which balances improved speed with minimal performance impact.

\begin{table*}[htbp]
\centering
  \footnotesize
  \caption{Results of different partial reconstruction strategies. Reconstruction Ratio=25\%. Top-1 classification accuracy is reported. All pretrained on the MillionAID dataset.
  }
    \resizebox{0.7\linewidth}{!}{\begin{tabular}{cccc}
   \hline
    \multirow{2}{*}{Method} & \multirow{2}{*}{Selection} &  RESISC-45 & AID \\
    &  & OA (TR=20\% / 50\%)& OA (TR=20\% / 50\%)\\
    \hline
    Partial Recognition & False  &  89.2/93.6  & 94.9/97.4\\\hline
    Partial Recognition  & Random  &  88.8/92.9  & 94.1/96.8\\
    Partial Recognition  & ATS~\cite{adamae}  & 85.7/87.4 & 88.9/91.7\\
    \cellcolor{black!15}Partial Recognition  & \cellcolor{black!15} HOG & \cellcolor{black!15}89.2/93.4 &  \cellcolor{black!15}94.8/97.2 \\
    \hline
    \end{tabular}}
    \label{table-ablation1}
\end{table*}

\begin{table*}[htp]
    \caption{Ablation study on the design choices of SelectiveMAE with ViT-B backbone pre-trained on MillionAID for 1,600 epochs. We report the top-1 fine-tuning accuracy ($\%$) on the RESISC-45. The default settings of SelectiveMAE are highlighted in grey.}
    \centering
    \tiny 
    \subfloat[
    Reconstruction ratio. 
    \label{reco_ratio}]{
        \begin{minipage}{0.455\linewidth}
            \centering   
            \resizebox{\linewidth}{!}{
            \begin{tabular}{ccc}
            \toprule
                \multirow{2}{*}{Reco. Ratio} & AID& RESISC-45   \\ \cmidrule(lr){2-3}
               &   \tiny{OA (TR=20\% / 50\%)}  & \tiny{OA (TR=10\% / 20\%)}   \\
                \midrule
                15\% & 94.80/96.97  &   90.24/93.55   \\
                \cellcolor{black!15}25\%  &
                \cellcolor{black!15}95.41/97.92 & \cellcolor{black!15}91.32/94.12 \\
                \bottomrule
            \end{tabular}
            }
        \end{minipage}}
 \subfloat[
    Decoder depth.
    \label{decoder_depth}]{
        \begin{minipage}{0.48\linewidth}
            \centering
            \resizebox{\linewidth}{!}{
            \begin{tabular}{ccc}
            \toprule
                \multirow{2}{*}{Deocder Depth}  & AID& RESISC-45   \\ \cmidrule(lr){2-3}
               & \tiny{OA (TR=20\% / 50\%)}  & \tiny{OA (TR=10\% / 20\%)}  \\  
                \midrule
                4    & 95.16/97.58 &91.03/93.84 \\
                \cellcolor{black!15}12      & \cellcolor{black!15}95.41/97.92 & \cellcolor{black!15}91.32/94.12\\\bottomrule
            \end{tabular}
            }
        \end{minipage}}
    \\
        \vspace{1em}
        \subfloat[
    Mask ratio.
    \label{mask_ratio}]{
        \begin{minipage}{0.44\linewidth}
            \centering
            \resizebox{\linewidth}{!}{
            \begin{tabular}{ccc}
            \toprule
                \multirow{2}{*}{Mask Ratio} & AID& RESISC-45   \\ \cmidrule(lr){2-3}
              &  \tiny{OA (TR=20\% / 50\%)}  & \tiny{OA (TR=10\% / 20\%)}   \\
                \midrule
                95\%  &93.15/96.08 &89.48/90.63 \\
                \cellcolor{black!15}85\% & \cellcolor{black!15}95.41/97.92 & \cellcolor{black!15}91.32/94.12 \\\bottomrule
            \end{tabular}
            }
        \end{minipage}}
        \vspace{1em}
    \subfloat[
    Selection strategy in PSTS.
    \label{smooth}]{
        \begin{minipage}{0.48\linewidth}
            \centering
            \resizebox{\linewidth}{!}{
            \begin{tabular}{ccc}
            \toprule
               \multirow{2}{*}{Selection Strategy}  & AID& RESISC-45   \\ \cmidrule(lr){2-3}
               & \tiny{OA (TR=20\% / 50\%)}  & \tiny{OA (TR=10\% / 20\%)}  \\
                \midrule
            far-near-random & 93.24/96.18  & 88.41/92.10 \\
            \cellcolor{black!15}near-far-random  & \cellcolor{black!15}95.41/97.92 & \cellcolor{black!15}91.32/94.12 \\\bottomrule
            \end{tabular}
            }
        \end{minipage}}
    \label{table-ablation2}
\end{table*}

Table \ref{table-ablation2} summarizes the ablation experiments investigating various design choices of the proposed SelectiveMAE method, including Reconstruction Ratio, Mask Ratio, Decoder Depth, and the selection strategy in PSTS.

\textbf{(a) Reconstruction Ratio.} We explored different reconstruction ratios for SelectiveMAE. Reducing the reconstruction ratio to 15\% resulted in significant performance degradation in downstream tasks. We found that a 25\% reconstruction ratio strikes a balance between speed and performance in SelectiveMAE.

\textbf{(b) Decoder Depth.} Decreasing the number of decoder layers notably reduced performance. Given SelectiveMAE's emphasis on patch selection, we maintained 12 decoder layers without modification, providing a stable baseline for future enhancements.

\textbf{(c) Mask Ratio.} Adjusting the mask ratio affected the number of input patches to the encoder and thus performance. A lower mask ratio improved performance by increasing the input patches, whereas a higher ratio accelerated processing, particularly with larger encoders. In our experiments, using ViT-L yielded a \textbf{2.3$\times$} acceleration compared to \textbf{2.1$\times$} with ViT-B, attributable to a higher mask ratio. Hence, an 85\% mask ratio was identified as the optimal balance between speed and accuracy.

\textbf{(d) Selection Strategy in Progressive Semantic Token Selection (PSTS).} In SelectiveMAE, PSTS begins by learning tokens that are close in distance, which provides consistent and easier patches for reconstruction. As training progresses, PSTS moves on to tokens that are farther apart, offering complementary and more challenging samples. Finally, PSTS randomly selects tokens to enhance model robustness. When we changed the selection strategy to far-near-random, training was hampered by frequent gradient explosions during pre-training. To mitigate this, we reduced the learning rate to 3e-5 (one-fifth of the original) and the batch size to 512. Although these adjustments allowed us to complete the far-near-random training, performance significantly declined.

While we have explored various design choices for SelectiveMAE, we believe its potential as a baseline method warrants further investigation. Future work may unlock greater improvements in both speed and accuracy.

\subsection{Configurations of Pre-training and Fine-tuning}
\label{app-sec3}
This section presents the datasets and implementation details for both pre-training and fine-tuning.

\textbf{Pre-training}: The default settings, detailed in Table~\ref{table-settings} (\texttt{i}), follow the official MAE implementation. We scale the learning rate according to the ratio of the mask ratio ($m$) to the reconstruction ratio ($r$) to match the loss variance of MAE. We use 12 decoder blocks with an 85\% mask ratio and a 25\% reconstruction ratio. For the 800-epoch experiments, the warm-up period is adjusted to 60 epochs. All other hyperparameters remain the same as in MAE.

\begin{table*}[htbp]
\centering
   \begin{subtable}[t]{0.6\linewidth}
        \centering
        \setlength\tabcolsep{2.5pt}%
        \small
        \resizebox{\linewidth}{!}{
        \begin{tabular}{l|c|cc}
        \hline
        Task                    & \multicolumn{1}{c|}{\texttt{(i)} Pre-training}                                                                                                                                                                                                                                   & \multicolumn{2}{c}{\texttt{(ii)} Scene Classification}                                                                                                                                                                            \\ \hline
        Dataset                  & OpticalRS-13M  & AID                                                                 & RESISC-45                                           \\ \hline
        Optimizer               & AdamW                                                               & AdamW                                &   AdamW       \\
        Input Size              & 224$\times$224                                                             & 224$\times$224                       & 224$\times$224                                                         \\
        Input channel           & RGB                                                                 & RGB                  & RGB       \\
        Base learning rate      &1.5e-4 & 1e-3                                                         & 1e-3                                     \\
        Learning rate scheduler & \begin{tabular}[c]{@{}c@{}}Cosine\\ Annealing\end{tabular}          & \begin{tabular}[c]{@{}c@{}}Cosine\\ Annealing\end{tabular}      & \begin{tabular}[c]{@{}c@{}}Cosine\\ Annealing\end{tabular}                                                 \\
        Weight decay            & 0.05                                                                & 0.05                                & 0.05                  \\
        Optimizer momentum       & (0.9, 0.95)  & (0.9, 0.999)                                                        & (0.9, 0.999)                                                                \\
        Batch size              &1024   & 64                                                                  & 64                                        \\
        Max iteration/epoch     &800 epoch  & 200 epoch                                                           & 200 epoch                                   \\
        Warmup                  & linear                                                              & linear                           & linear                           \\
        Warmup iteration/epoch   & 60 epoch & 5 epoch                                                             & 5 epoch                                              \\
        Drop path rate           & -      & 0.1                                                                 & 0.1                                       \\
        Augmentation            & \begin{tabular}[c]{@{}c@{}}RandoCrop,\\ RandomFlip\end{tabular}   & \begin{tabular}[c]{@{}c@{}}RandomCrop,\\ RandomErasing\end{tabular} & \begin{tabular}[c]{@{}c@{}}RandomCrop,\\ RandomErasing\end{tabular}                                                         \\
        Head/Detector          & -    & \begin{tabular}[c]{@{}c@{}}Linear\\ Classifier\end{tabular}             & \begin{tabular}[c]{@{}c@{}}Linear\\ Classifier\end{tabular}         \\
        Loss function               &-      & CrossEntropy                                                             & CrossEntropy                                  \\\hline
        \end{tabular}
        }
        \vspace{4mm}
       
    \end{subtable}
        \begin{subtable}[t]{0.8\linewidth}
        \centering
        
        \small
         \resizebox{\linewidth}{!}{
        \begin{tabular}{l|cc|cc}
        
        \hline
        Task     & \multicolumn{2}{c|}{\texttt{(iii)} Semantic Segmentation}   & \multicolumn{2}{c}{\texttt{(iv)} Object Detection}  \\ \hline
        Dataset         & LoveDA  & SpaceNetv1 & DIOR         &DIOR-R         \\ \hline
        Optimizer     & AdamW     & AdamW     & AdamW        & AdamW    \\
        Input Size         & 512 $\times$ 512   & 384 $\times$ 384       & 800$\times$800      & 800$\times$800         \\
        Input channel      & RGB         & RGB     & RGB     & RGB        \\
        Base learning rate         & 6e-5         & 6e-5      & 1e-4         & 1e-4      \\
        Learning rate scheduler      &\begin{tabular}[c]{@{}c@{}}Cosine\\ Annealing\end{tabular}     & \begin{tabular}[c]{@{}c@{}}Cosine\\ Annealing\end{tabular}    & Multistep         & Multistep     \\
        Weight decay      & 0.05      & 0.05     & 0.05         & 0.05        \\

        Batch size         & 8         & 8      & 4       & 4        \\
        Max iteration/epoch     & 80k iters      & 80k iters       & 12 epoch     & 12 epoch     \\
        Warmup      & linear      & linear        & linear       & linear     \\
        Warmup iteration/epoch     & 1.5k iters      & 1.5k iters      & 0.5k Iters     & 0.5k iters    \\
        Warmup ratio        & 1e-6      & 1e-6       & 1e-6         & 1e-6       \\
        Drop path rate      & 0.1         & 0.1        & 0.1     & 0.1     \\
        Augmentation     & \begin{tabular}[c]{@{}c@{}}RandomScaling \\ (0.5 to 2.0),\\ RandomCrop,\\ RandomFlip\end{tabular} & \begin{tabular}[c]{@{}c@{}}RandomScaling \\ (0.5 to 2.0),\\ RandomCrop,\\ RandomFlip\end{tabular} & RandomFlip   & RandomFlip    \\
        Head/Detector         & UperNet     & UperNet     & Faster-RCNN     & Oriented-RCNN        \\
        Loss function       & CrossEntropy       & CrossEntropy         & \begin{tabular}[c]{@{}c@{}}CrossEntropy,\\ L1 \end{tabular}     & \begin{tabular}[c]{@{}c@{}}CrossEntropy,\\ SmoothL1 \end{tabular}         \\\hline
        \end{tabular}
        }
    \end{subtable}

    \caption{Detailed configurations of pre-training and fine-tuning.}
    \label{table-settings}
\end{table*}

\textbf{Scene Classification.} We conducted scene classification experiments using a standard linear classifier on two commonly used datasets: AID and NWPU-RESISC45. Implementation details are summarized in Table \ref{table-settings} (\texttt{ii}).

\begin{enumerate}[label=\arabic*)]
\item \textit{AID.}
This dataset contains 10,000 images, each sized 600$\times$600 pixels with a Ground Sample Distance (GSD) ranging from 0.5 to 8 meters. The images are categorized into 30 classes, each with approximately 220 to 400 images. We follow standard protocols in RVSA~\cite{RVSA}, using $x\%$ of the data for training and the remaining $(1-x)\%$ for testing, where $x \in \{20, 50\}$.

\item \textit{NWPU-RESISC45 (RESISC-45).} 
This dataset comprises 31,500 images, each sized 256$\times$256 pixels with a GSD ranging from 0.5 to 30 meters. It is divided into 45 categories, each containing 700 images. We use two settings, \textit{i.e.}, 10\% (and 20\%) of the data for training, with the remaining 90\% (and 80\%) for testing, in line with previous works.
\end{enumerate}

\noindent\textbf{Semantic Segmentation.} 
Semantic segmentation is extensively studied in remote sensing, which aims to automate the extraction of land use classes and ground instances. For this experiment, considering factors such as spatial resolution, spectrum, and the number of categories, we chose two well-known datasets:
\begin{enumerate}[label=\arabic*)]
\item\textit{LoveDA.} 
This dataset includes urban and rural scenes with 0.3m resolution imagery from Google Earth, captured in July 2016, covering 536.15 km² across Nanjing, Changzhou, and Wuhan. It consists of 5,987 images, each 1,024$\times$1,024 pixels, and includes seven common land cover types. We combined the official training and validation sets for training and used the official testing set for evaluation, following the common practice.
\item\textit{SpaceNetv1.} 
Provided by the SpaceNet Challenge, this dataset is intended for extracting building footprints. It includes DigitalGlobe WorldView-2 satellite imagery with a 0.5m GSD, captured from 2011 to 2014, covering approximately 2,544 km² over Rio de Janeiro. It contains 382,534 building instances. We used the 6,940 images from the original training set, randomly splitting them into 5,000 images for training and the remainder for testing, in line with previous studies.
\end{enumerate}
We utilized UperNet as the segmentation head based on MMSegmentation\footnote{\url{https://github.com/open-mmlab/mmsegmentation}}, as described in \cite{RVSA,MTP}. Detailed fine-tuning settings are provided in Table \ref{table-settings} (\texttt{iii}).

\noindent\textbf{Horizontal \& Oriented Objection Detection.}
We use the DIOR dataset to assess the performance of SelectiveMAE and other RSFMs in horizontal object detection tasks. Following RVSA~\cite{RVSA}, we employ Faster-RCNN as the detector, as detailed in Table \ref{table-settings} (\texttt{iv}).
\begin{enumerate}[label=\arabic*)]
\item \textit{DIOR.} 
This dataset consists of 23,463 visible remote sensing images with 192,472 object instances, annotated with horizontal bounding boxes across 20 common object classes. Each image of size 800$\times$800 has a GSD ranging from 0.5 to 30 meters. The dataset is split into 5,862 training patches, 5,863 validation patches, and 11,738 test patches. Following RVSA~\cite{RVSA}, we merge the training and validation sets for training, using the test set for evaluation. The high inter-class similarity and intra-class diversity pose significant challenges for precise localization and classification.
\end{enumerate}
Remote sensing images include diverse objects such as buildings, vehicles, and bridges, which are densely distributed and vary in size, scale, and orientation. This makes object detection particularly challenging, especially for oriented object detection. To evaluate RSFMs on this task, we use the DIOR-R dataset and Oriented-RCNN as the detector, as detailed in Table \ref{table-settings} (\texttt{iv}).
\begin{enumerate}[label=\arabic*)]
\setcounter{enumi}{1}
\item \textit{DIOR-R.} 
This dataset uses the same images as DIOR but includes oriented bounding boxes, making it suitable for oriented object detection. Following RVSA~\cite{RVSA}, we combine the training and validation sets for training, using the test set for evaluation.
\end{enumerate}

For horizontal object detection and oriented object detection, we use MMDetection\footnote{\url{https://github.com/open-mmlab/mmdetection}} and MMRotate\footnote{\url{https://github.com/open-mmlab/mmrotate}} for implementation, respectively.

\subsection{Qualitative Results}
\label{app-vis}

\subsubsection{Visualization of OpticalRS-13M Categories and Samples.}
\begin{figure*}[htbp]
    \centering
    \includegraphics[width=0.9\linewidth]{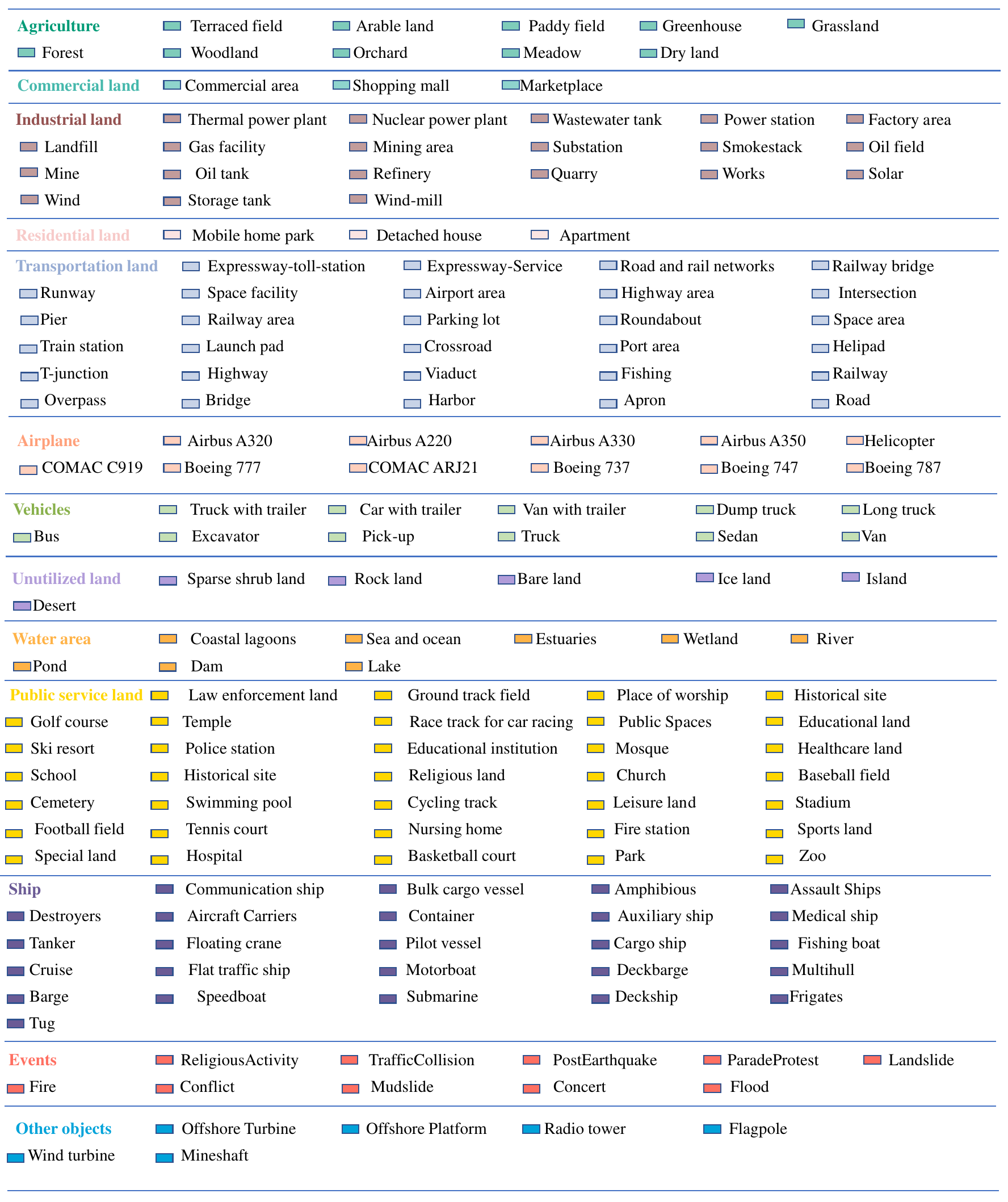}
    \caption{Category structure of OpticalRS-13M.
}
    \label{fig:OpticalRS-13M-categories}
\end{figure*}

\begin{figure*}[htbp]
    \centering
    \includegraphics[width=0.8\linewidth]{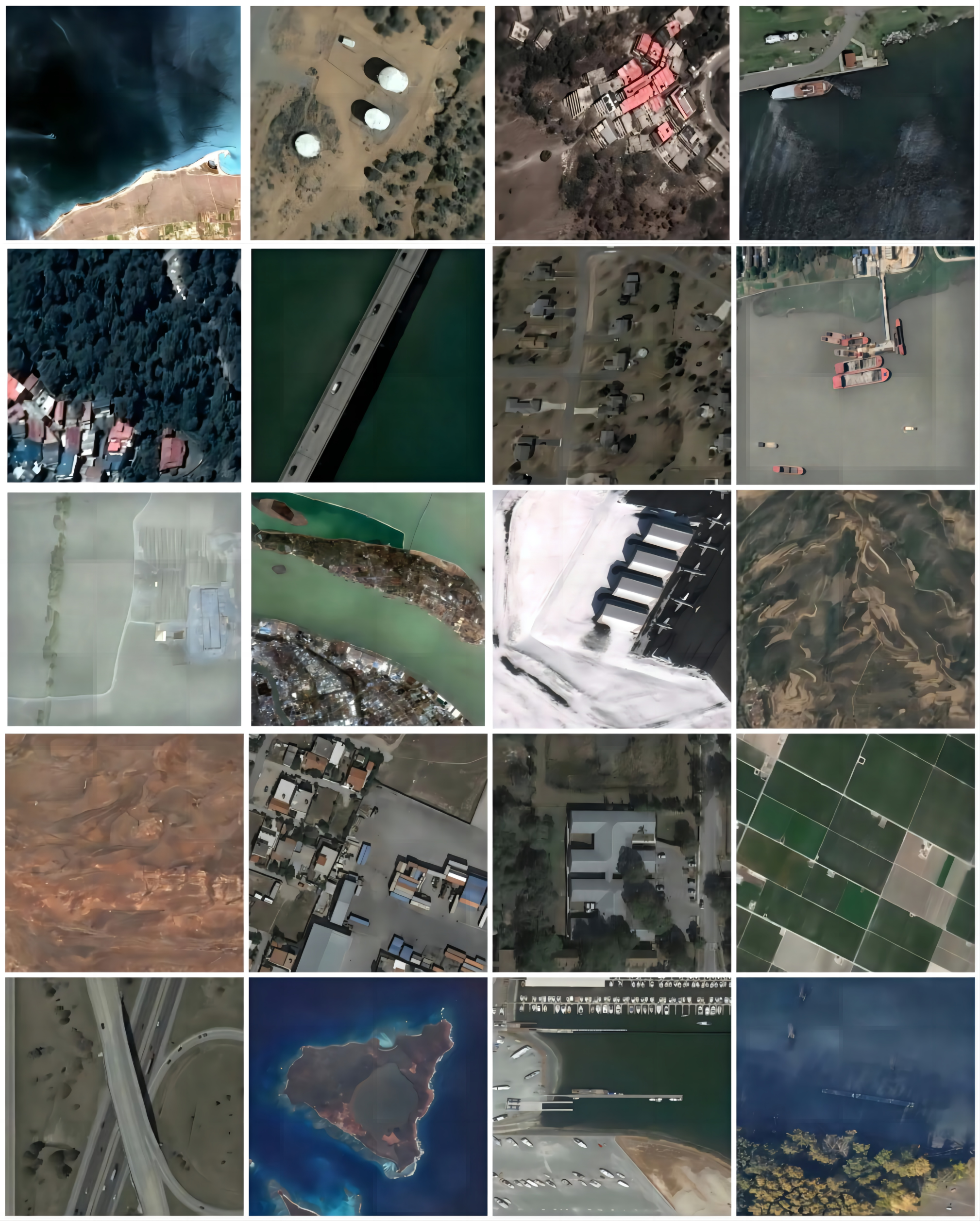}
    \caption{Visualization of OpticalRS-13M Samples.}
    \label{fig:OpticalRS-13M}
\end{figure*}

To further highlight the diversity of OpticalRS-13M, we present its category structure in Figure \ref{fig:OpticalRS-13M-categories}. It can be seen that, OpticalRS-13M encompasses a wide range of diverse RS scenarios encountered in downstream tasks such as object-level detection and pixel-level segmentation. Meanwhile, the diverse data contained in OpticalRS-13M also provides finer grained detail information to support various downstream tasks.

We selected images from the OpticalRS-13M dataset, as displayed in Figure \ref{fig:OpticalRS-13M}. It can be seen that, OpticalRS-13M encompasses a wide range of diverse RS scenarios and provides finer grained detail information to support various downstream tasks.  Nonetheless, we also find that optical remote sensing images contain numerous redundant background pixels, and the high-value information in these images often occupies only a small portion of the total pixel, a characteristic particularly evident in downstream tasks like identifying ships or bridges in satellite images.

\subsubsection{Visualization of SelectiveMAE Results on Downstream Tasks}
\begin{figure*}[htbp]
    \centering
    \includegraphics[width=0.8\linewidth]{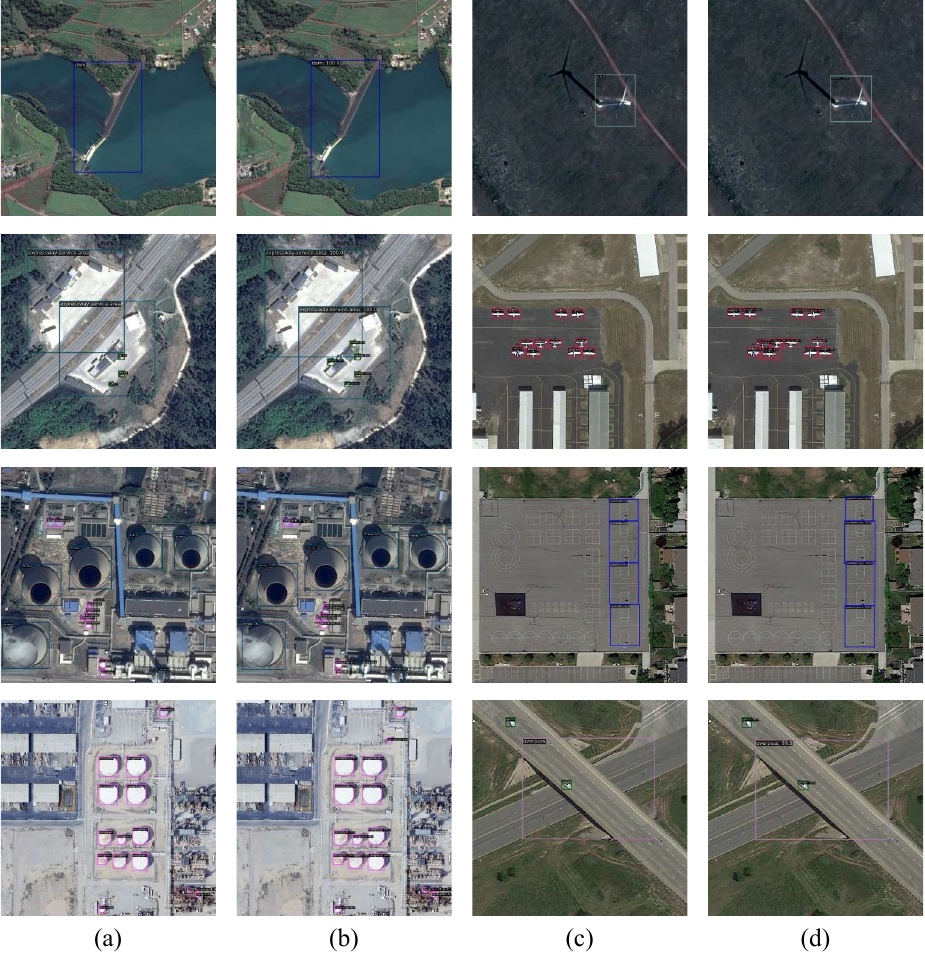}
    \caption{Visualization of SelectiveMAE predictions on the DIOR testing set. (a)(c) Ground truth. (b)(d) Predicted results of SelectiveMAE.}
    \label{fig:dior}
\end{figure*}

\begin{figure*}[htbp]
    \centering
    \includegraphics[width=0.8\linewidth]{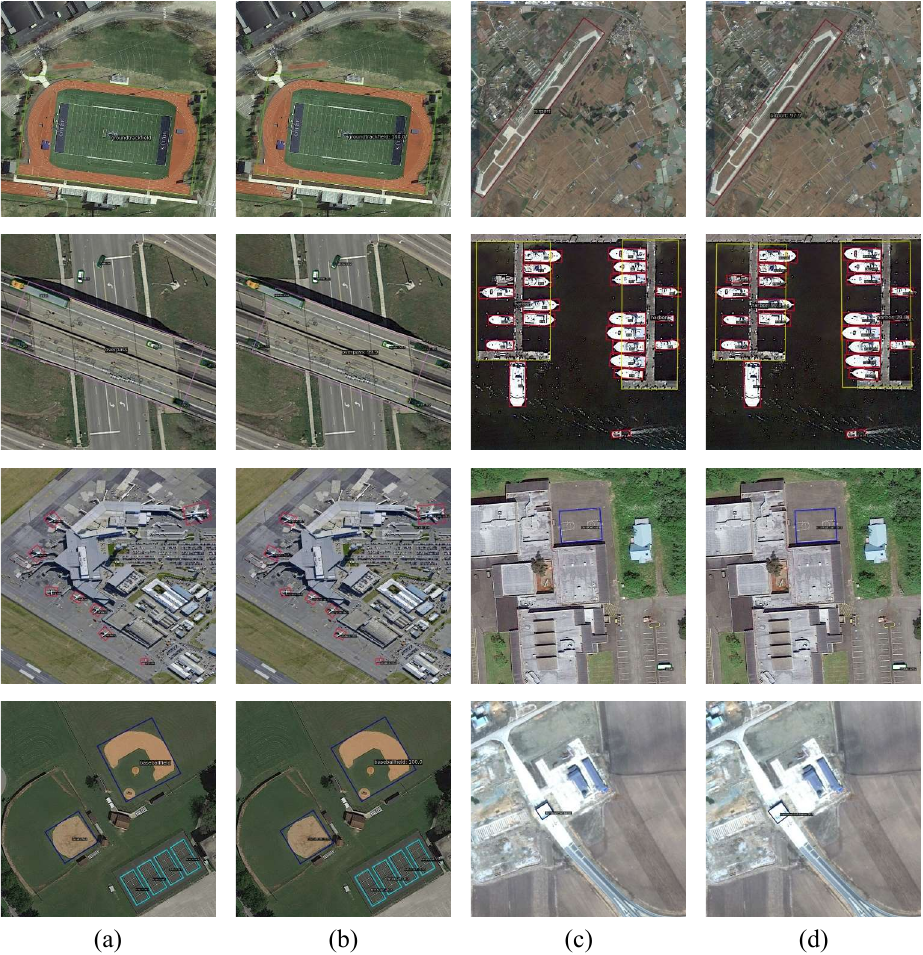}
    \caption{Visualization of SelectiveMAE predictions on the DIOR-R testing set. (a)(c) Ground truth. (b)(d) Predicted results of SelectiveMAE.}
    \label{fig:diorr}
\end{figure*}

\begin{figure*}[htbp]
    \centering
    \includegraphics[width=0.8\linewidth]{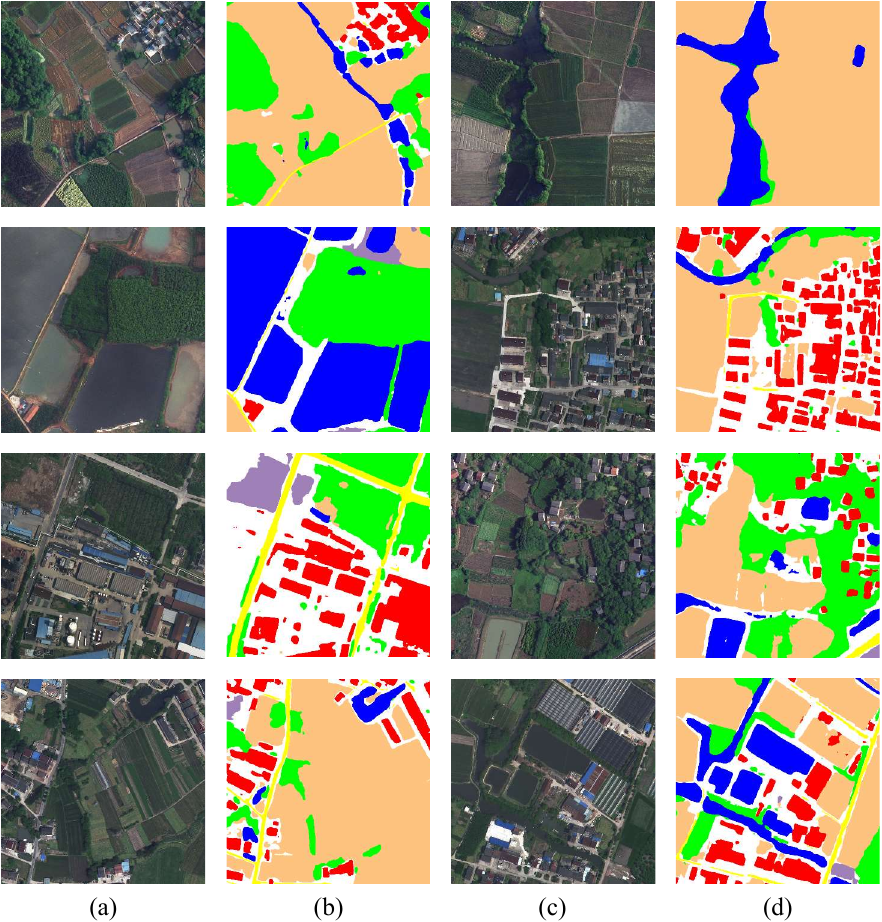}
    \caption{
    Visualization of SelectiveMAE predictions on the LoveDA testing set. (a)(c) Testing images. (b)(d) Predicted results of SelectiveMAE.
    }
    \label{fig:loveda}
\end{figure*}

\begin{figure*}[htbp]
    \centering
    \includegraphics[width=0.8\linewidth]{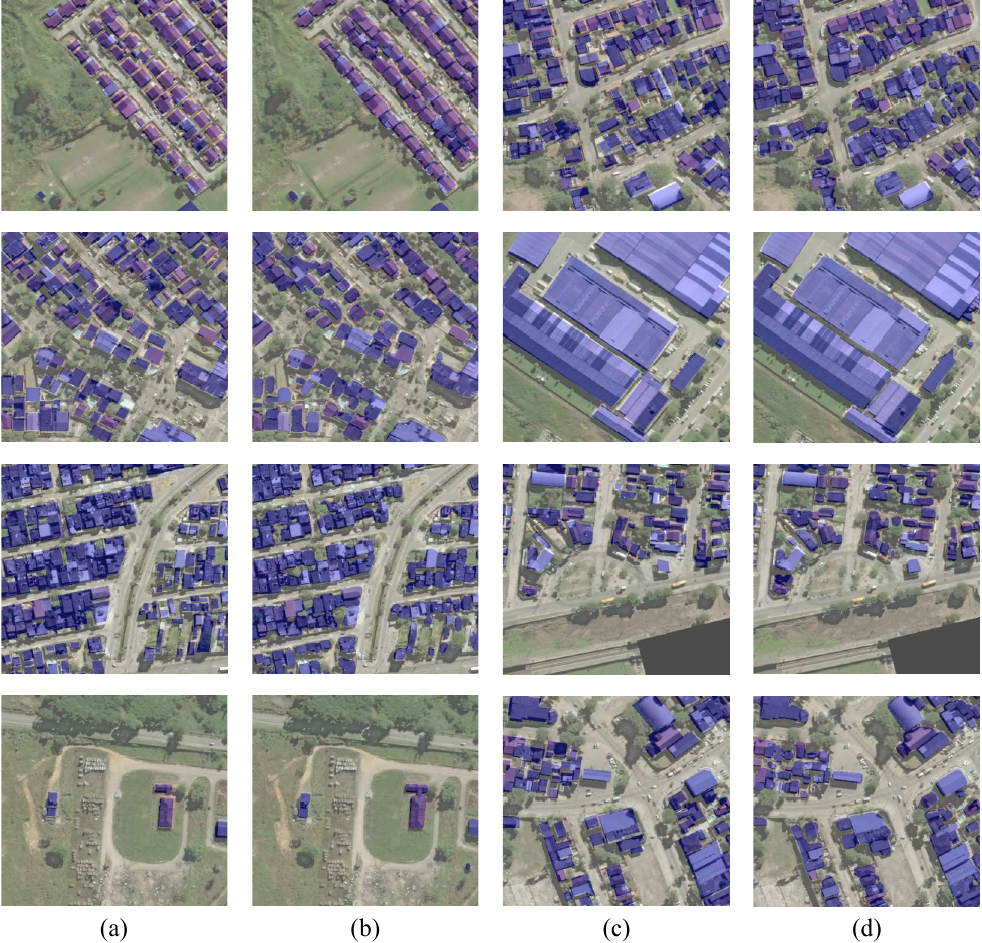}
    \caption{
    Visualization of SelectiveMAE predictions on the SpaceNetv1 testing set. (a)(c) Ground truth. (b)(d) Predicted results of SelectiveMAE.
    }
    \label{fig:spacenetv1}
\end{figure*}

Figures \ref{fig:dior}-\ref{fig:spacenetv1} present the results of our SelectiveMAE on the DIOR, DIOR-R, LoveDA, and SpaceNetv1 datasets. We utilized the ViT-L pre-trained on OpticalRS-13M as the backbone. The results closely match the ground truth, ensuring high accuracy. For detection tasks, our methods accurately identify diverse objects of various sizes. In segmentation tasks, they facilitate extensive extraction and mapping of significant RS land cover categories. In summary, OpticalRS-13M and SelectiveMAE enable the successful construction of effective RS foundation models.

\subsection{Datasheets}
\label{app-datasheets}

In this section, we follow the NeurIPS Dataset and Benchmark guideline and use the template from Gebru \textit{et al.} \cite{datasheets} to document necessary information about the proposed datasets and benchmarks.

\subsubsection{Motivation}
The questions in this section are primarily intended to encourage dataset creators to clearly articulate their reasons for creating the dataset and to promote transparency about funding interests. The latter may be particularly relevant for datasets created for research purposes.
\begin{enumerate}
    \item \textit{``For what purpose was the dataset created?''}
    
    \textcolor{BurntOrange}{\textbf{A:}} The dataset was created to support research on remote sensing foundation models (RSFMs) using self-supervised learning techniques.
    
    \item \textit{``Who created the dataset (\textit{e.g.}, which team, research group) and on behalf of which entity?''}
    
    \textcolor{BurntOrange}{\textbf{A:}} The dataset was created by: Anonymous authors.

    \item \textit{``Who funded the creation of the dataset?''}
    
    \textcolor{BurntOrange}{\textbf{A:}}
    The dataset creation was funded by the affiliations of the authors involved in this work.
\end{enumerate}

\subsubsection{Composition}
Most of the questions in this section are intended to provide dataset consumers with the information they need to make informed decisions about using the dataset for their chosen tasks. Some of the questions are designed to elicit information about compliance with the EU’s General Data Protection Regulation (GDPR) or comparable regulations in other jurisdictions. Questions that apply only to datasets that relate to people are grouped together at the end of the section. We recommend taking a broad interpretation of whether a dataset relates to people. For example, any dataset containing text that was written by people relates to people.
\begin{enumerate}
    \item \textit{``What do the instances that comprise our datasets represent (\textit{e.g.}, documents, photos, people, countries)?''}
    
    \textcolor{BurntOrange}{\textbf{A:}} The dataset primarily comprises visible light remote sensing images captured by satellites. All datasets utilized in OpticalRS-13M are publicly accessible and nonprofit.
    
    \item \textit{``How many instances are there in total (of each type, if appropriate)?''}
    
    \textcolor{BurntOrange}{\textbf{A:}} OpticalRS-13M contains 13 million remote sensing image instances captured by satellites.

    \item \textit{``Does the dataset contain all possible instances or is it a sample (not necessarily random) of instances from a larger set?''}
    
    \textcolor{BurntOrange}{\textbf{A:}} Yes, our dataset contains all possible instances that have been collected so far.
    
    \item \textit{``Is there a label or target associated with each instance?''}
    
    \textcolor{BurntOrange}{\textbf{A:}} No, our dataset is intended for self-supervised learning. Therefore, each instance is an individual remote sensing image and does not contain annotations.
    
    \item \textit{``Is any information missing from individual instances?''}
    
    \textcolor{BurntOrange}{\textbf{A:}} No.
    
    \item \textit{``Are relationships between individual instances made explicit (\textit{e.g.}, users’ movie ratings, social network links)?''}
    
    \textcolor{BurntOrange}{\textbf{A:}} Yes, the relationship between individual instances is explicit.
    
    \item \textit{``Are there recommended data splits (\textit{e.g.}, training, development/validation, testing)?''}
    
    \textcolor{BurntOrange}{\textbf{A:}} Yes, the entire dataset is intended for self-supervised methods, and we recommend using the whole dataset for self-supervised learning research.
    
    \item \textit{``Is the dataset self-contained, or does it link to or otherwise rely on external resources (\textit{e.g.}, websites, tweets, other datasets)?''}
    
    \textcolor{BurntOrange}{\textbf{A:}} Yes, our dataset relies on many publicly available remote sensing datasets, which we have detailed in the main text.
    
    \item \textit{``Does the dataset contain data that might be considered confidential (\textit{e.g.}, data that is protected by legal privilege or by doctor–patient confidentiality, data that includes the content of individuals’ non-public communications)?''}
    
    \textcolor{BurntOrange}{\textbf{A:}} No, all data are clearly licensed.
    
    \item \textit{``Does the dataset contain data that, if viewed directly, might be offensive, insulting, threatening, or might otherwise cause anxiety?''}
    
    \textcolor{BurntOrange}{\textbf{A:}}
    No.
\end{enumerate}

\subsubsection{Collection Process}
In addition to the goals outlined in the previous section, the questions in this section are designed to elicit information that may help researchers and practitioners create alternative datasets with similar characteristics. Again, questions that apply only to datasets that relate to people are grouped together at the end of the section.
\begin{enumerate}
    \item \textit{``How was the data associated with each instance acquired?''}
    
    \textcolor{BurntOrange}{\textbf{A:}} Please refer to the details listed in the main text Sec. 3.
    
    \item \textit{``What mechanisms or procedures were used to collect the data (\textit{e.g.}, hardware apparatuses or sensors, manual human curation, software programs, software APIs)?''}
    
    \textcolor{BurntOrange}{\textbf{A:}} Please refer to the details listed in the main text Sec. 3.
    
    \item \textit{``If the dataset is a sample from a larger set, what was the sampling strategy (\textit{e.g.}, deterministic, probabilistic with specific sampling probabilities)?''} 
    
    \textcolor{BurntOrange}{\textbf{A:}} Please refer to the details listed in the main text Sec. 3.
\end{enumerate}

\subsubsection{Preprocessing, Cleaning, and Labeling}
The questions in this section are intended to provide dataset
consumers with the information they need to determine whether the “raw” data has been processed in ways that are compatible with their chosen tasks. For example, text that has been converted into a ``bag-of-words" is not suitable for tasks involving word order.
\begin{enumerate}
    \item \textit{``Was any preprocessing/cleaning/labeling of the data done (\textit{e.g.}, discretization or bucketing, tokenization, part-of-speech tagging, SIFT feature extraction, removal of instances, processing of missing values)?''}
    
    \textcolor{BurntOrange}{\textbf{A:}} Yes, we preprocessed and cleaned data in our dataset.

    \item \textit{``Was the `raw' data saved in addition to the preprocessed/cleaned/labeled data (\textit{e.g.}, to support unanticipated future uses)?''} 
    
    \textcolor{BurntOrange}{\textbf{A:}} Yes, raw data is accessible.
    
    \item \textit{``Is the software that was used to preprocess/clean/label the data available?''} 
    
    \textcolor{BurntOrange}{\textbf{A:}} Yes, the necessary software used to preprocess and clean the data is publicly available.
\end{enumerate}

\subsubsection{Uses}
The questions in this section are intended to encourage dataset creators to reflect on tasks for which the dataset should and should not be used. By explicitly highlighting these tasks, dataset creators can help dataset consumers make informed decisions, thereby avoiding potential risks or harms.
\begin{enumerate}
    \item \textit{``Has the dataset been used for any tasks already?''} 
    
    \textcolor{BurntOrange}{\textbf{A:}} No.
    
    \item \textit{``Is there a repository that links to any or all papers or systems that use the dataset?''} 
    
    \textcolor{BurntOrange}{\textbf{A:}} Yes, we provide such links in our GitHub repository.
    
    \item \textit{``What (other) tasks could the dataset be used for?''} 
    
    \textcolor{BurntOrange}{\textbf{A:}} The dataset could be used for training the remote sensing foundation models with the self-supervised learning method.
    
    \item \textit{``Is there anything about the composition of the dataset or the way it was collected and preprocessed/cleaned/labeled that might impact future uses?''} 
    
    \textcolor{BurntOrange}{\textbf{A:}} N/A.
    
    \item \textit{``Are there tasks for which the dataset should not be used?''} 
    
    \textcolor{BurntOrange}{\textbf{A:}} N/A.
\end{enumerate}

\subsubsection{Distribution}
Dataset creators should provide answers to these questions prior to distributing the dataset either internally within the entity on behalf of which the dataset was created or externally to third parties.
\begin{enumerate}
    \item \textit{``Will the dataset be distributed to third parties outside of the entity (\textit{e.g.}, company, institution, organization) on behalf of which the dataset was created?''} 
    
    \textcolor{BurntOrange}{\textbf{A:}} No.
    
    \item \textit{``How will the dataset be distributed (\textit{e.g.}, tarball on website, API, GitHub)?''} 
    
    \textcolor{BurntOrange}{\textbf{A:}} Very likely to be distributed by website, API, and GitHub repository.
    
    \item \textit{``When will the dataset be distributed?''} 
    
    \textcolor{BurntOrange}{\textbf{A:}} The datasets are publicly accessible.
    
    \item \textit{``Will the dataset be distributed under a copyright or other intellectual property (IP) license, and/or under applicable terms of use (ToU)?''} 
    
    \textcolor{BurntOrange}{\textbf{A:}} Yes, the dataset is under the Creative Commons Attribution-NonCommercial-ShareAlike 4.0 International License.
    
    \item \textit{``Have any third parties imposed IP-based or other restrictions on the data associated with the instances?''} 
    
    \textcolor{BurntOrange}{\textbf{A:}} No.
    
    \item \textit{``Do any export controls or other regulatory restrictions apply to the dataset or to individual instances?''} 
    
    \textcolor{BurntOrange}{\textbf{A:}} No.    
\end{enumerate}

\subsubsection{Maintenance}
As with the questions in the previous section, dataset creators should provide answers to these questions prior to distributing the dataset. The questions in this section are intended to encourage dataset creators to plan for dataset maintenance and communicate this plan to dataset consumers.
\begin{enumerate}
    \item \textit{``Who will be supporting/hosting/maintaining the dataset?''} 
    
    \textcolor{BurntOrange}{\textbf{A:}} The authors of this work serve to support, host, and maintain the datasets.
    
    \item \textit{``How can the owner/curator/manager of the dataset be contacted (\textit{e.g.}, email address)?''} 
    
    \textcolor{BurntOrange}{\textbf{A:}} The curators can be contacted via the email addresses listed on our webpage.
    
    \item \textit{``Is there an erratum?''} 
    
    \textcolor{BurntOrange}{\textbf{A:}} There is no explicit erratum; updates and known errors will be specified in future versions.
    
    \item \textit{``Will the dataset be updated (\textit{e.g.}, to correct labeling errors, add new instances, delete instances)?''} 
    
    \textcolor{BurntOrange}{\textbf{A:}} Yes, for the current version. Future updates (if any) will be posted on the dataset website.
    
    \item \textit{``Will older versions of the dataset continue to be supported/hosted/maintained?''} 
    
    \textcolor{BurntOrange}{\textbf{A:}} Yes. This is the first version of the release; future updates will be posted and older versions will be replaced.
    
    \item \textit{``If others want to extend/augment/build on/contribute to the dataset, is there a mechanism for them to do so?''} 
    
    \textcolor{BurntOrange}{\textbf{A:}} Yes, we provide detailed instructions for future extensions.
\end{enumerate}

\subsection{Limitation and Potential Societal Impact}
\label{app-limitation}
In this section, we elaborate on the limitations and potential societal impact of this work.

\subsubsection{Potential Limitations}
While OpticalRS-13M provides a comprehensive benchmark for training the remote sensing foundation models with self-supervised learning methods, there are several limitations to consider:
\begin{itemize}
    \item \textbf{Scope of Sensors:} Although our benchmark includes 4 million visible light remote sensing images, it may not cover all possible real-world scenarios. There could be additional sensor data, like multispectral data that were not included in this study, potentially limiting the generalizability of our findings.

    \item \textbf{Model and Dataset Diversity:} While our dataset is primarily focused on adapting to downstream tasks like detection and segmentation, there is undeniably less data available for these tasks compared to general scene classification. In the future, we should collect more data that is better suited for various downstream tasks.

    \item \textbf{Computation and Resource Requirements:} Pre-training on the extensive OpticalRS-13M dataset, comprising 4 million images, demands substantial computational resources, despite our introduction of the efficient SelectiveMAE method. This may limit access to the benchmark for research groups lacking ample computational power.
\end{itemize}

\subsubsection{Potential Negative Societal Impact}
While the development of remote sensing foundation models with self-supervised learning methods has the potential to significantly advance remote sensing downstream tasks, there are potential negative societal impacts that must be considered:
\begin{itemize}
    \item \textbf{Safety Risks:} Our benchmark aims to improve remote sensing foundation models, but relying too heavily on these models may breed overconfidence in autonomous systems. It's crucial to deploy these systems with adequate safety measures and human oversight to uphold public safety.

    \item \textbf{Environmental Impact:} Training and evaluating remote sensing foundation models demand substantial computational resources, leading to a notable environmental footprint. Encouraging the adoption of energy-efficient algorithms (\textit{i.e.}, the proposed SelectiveMAE) and sustainable computing practices is crucial to reduce the environmental impact of this research.

    \item \textbf{Bias and Fairness:} The performance of foundational remote sensing models can vary across different environments and conditions, potentially introducing biases in downstream tasks within the remote sensing domain. It is essential to train and evaluate these models on diverse datasets to ensure fairness and avoid discriminatory outcomes.
\end{itemize}

\end{document}